\documentclass[sigconf]{acmart}

\usepackage[figurename=Fig.]{caption}
\usepackage[labelfont={rm}, textfont={rm}]{subcaption}
\usepackage{multirow}
\usepackage{enumitem}

\hypersetup{colorlinks=true,urlcolor=blue}

\setlist[itemize,1]{leftmargin=2em}

\setcopyright{none}
\acmDOI{10.1145/3159652.3159666}
\acmISBN{978-1-4503-5581-0/18/02}
\acmConference[WSDM'18]{the 11th ACM International Conference on Web Search and Data Mining}{February 5--9, 2018}{Marina Del Rey, CA, USA}

\acmYear{2018}
\copyrightyear{2018}
\acmPrice{15.00}

\begin{document}
\title[SHINE: Signed Heterogeneous Information Network Embedding]{SHINE: Signed Heterogeneous Information Network Embedding for Sentiment Link Prediction}

%\author[H. Wang et al.]{Hongwei Wang$^{1,2}$, Fuzheng Zhang$^2$, Min Hou$^{3, 2}$, Xing Xie$^2$, Minyi Guo$^1$, Qi Liu$^3$}
%\authornote{M. Guo is the corresponding author.}
%\affiliation{$^1$Shanghai Jiao Tong University, wanghongwei55@gmail.com, guo-my@cs.sjtu.edu.cn}
%\affiliation{$^2$Microsoft Research Asia, \{fuzzhang, xingx\}@microsoft.com}
%\affiliation{$^3$University of Science and Technology of China, hmhoumin@gmail.com, qiliuql@ustc.edu.cn}

\author{Hongwei Wang}
\affiliation{Shanghai Jiao Tong University\\Shanghai, China}
\email{wanghongwei55@gmail.com}
\authornote{This work is done while H. Wang and M. Hou are visiting Microsoft Research Asia.}

\author{Fuzheng Zhang}
\affiliation{Microsoft Research Asia\\Beijing, China}
\email{fuzzhang@microsoft.com}

\author{Min Hou}
\affiliation{University of Science and Technology of China, Hefei, Anhui, China}
\email{hmhoumin@gmail.com}

\author{Xing Xie}
\affiliation{Microsoft Research Asia\\Beijing, China}
\email{xingx@microsoft.com}

\author{Minyi Guo}
\affiliation{Shanghai Jiao Tong University\\Shanghai, China}
\email{guo-my@cs.sjtu.edu.cn}
\authornote{M. Guo is the corresponding author.}

\author{Qi Liu}
\affiliation{University of Science and Technology of China, Hefei, Anhui, China}
\email{qiliuql@ustc.edu.cn}

\begin{abstract}
	In online social networks people often express attitudes towards others, which forms massive sentiment links among users.
	Predicting the sign of sentiment links is a fundamental task in many areas such as personal advertising and public opinion analysis.
	Previous works mainly focus on textual sentiment classification, however, text information can only disclose the ``tip of the iceberg'' about users' true opinions, of which the most are unobserved but implied by other sources of information such as social relation and users' profile.
	To address this problem, in this paper we investigate how to predict possibly existing sentiment links in the presence of heterogeneous information.
	First, due to the lack of explicit sentiment links in mainstream social networks, we establish a labeled heterogeneous sentiment dataset which consists of users' sentiment relation, social relation and profile knowledge by entity-level sentiment extraction method.
	Then we propose a novel and flexible end-to-end \textit{Signed Heterogeneous Information Network Embedding} (SHINE) framework to extract users' latent representations from heterogeneous networks and predict the sign of unobserved sentiment links.
	SHINE utilizes multiple deep autoencoders to map each user into a low-dimension feature space while preserving the network structure.
	We demonstrate the superiority of SHINE over state-of-the-art baselines on link prediction and node recommendation in two real-world datasets.
	The experimental results also prove the efficacy of SHINE in cold start scenario.
\end{abstract}

%\keywords{sentiment link prediction; signed heterogeneous network embedding; online social networks}

\maketitle

\section{Introduction}
	The past decade has witnessed the proliferation of online social networks such as Facebook, Twitter and Weibo.
	In these social network sites, people often share feelings and express attitudes towards others, e.g., friends, movie stars or politicians, which forms \textit{sentiment links} among these users.
	Different from explicit social links indicating friend or follow relationship, sentiment links are implied by the semantic content posted by users, and involve different types: \textit{positive} sentiment links express like, trust or support attitudes, while \textit{negative} sentiment links signify dislike or disapproval of others.
	For example, a tweet saying ``\textit{Vote Trump!}'' shows a positive sentiment link from the poster to Donald Trump, and ``\textit{Trump is mad...}'' indicates the opposite case.
	
	For a given sentiment link, we define its \textit{sign} to be positive or negative depending on whether its related content expresses a positive or negative attitude from the generator of the link to the recipient \cite{leskovec2010predicting}, and all such sentiment links form a new network topology called \textit{sentiment network}.
	Previous work \cite{kiritchenko2014nrc, dos2014deep, nguyen2015phrasernn} mainly focuses on sentiment classification based on the concrete content posted by users.
	However, they cannot detect the existence of sentiment links without any prior content information, which greatly limits the number of possible sentiment links that could be found.
	For example, if a user does not post any word concerning Trump, it is impossible for traditional sentiment classifiers to extract the user's attitude towards him because ``one cannot make bricks without straw''.
	Therefore, a fundamental question is, can we predict the sign of a given sentiment link without observing its related content?
	%In other words, is there any other explicit or implicit pattern in sentiment network we can exploit to facilitate the prediction task?
	The solution to this problem will benefit a great many online services such as personalized advertising, new friends recommendation, public opinion analysis, opinion polls, etc.
	
	Despite the great importance, there is little prior work concerning predicting the sign of sentiment links among users in social networks.
	The challenges are two-fold.
	On the one hand, lack of explicit sentiment labels makes it difficult to determine the polarity of existing and potential sentiment links.
	On the other hand, the complexity of sentiment generation and the sparsity of sentiment links make it hard for algorithms to achieve desirable performance.
	Recently, several studies \cite{leskovec2010predicting, ye2013predicting, kumar2016edge, zheng2015spectral} propose methods to solve the problem of predicting signed links.
	However, they rely heavily on manually designed features and cannot work well in real-world scenarios.
	Another promising approach called \textit{network embedding} \cite{perozzi2014deepwalk, tang2015line, grover2016node2vec, wang2016structural}, which automatically learns features of users in network, seems plausible to solve the task.
	However, they can only apply to networks with positive-weighted (i.e., unsigned) and single-type (i.e., homogeneous) edges, which limits their power in the task of practical sentiment link prediction.
	
\begin{figure}[t]
		\centering
  		\includegraphics[width=.45\textwidth]{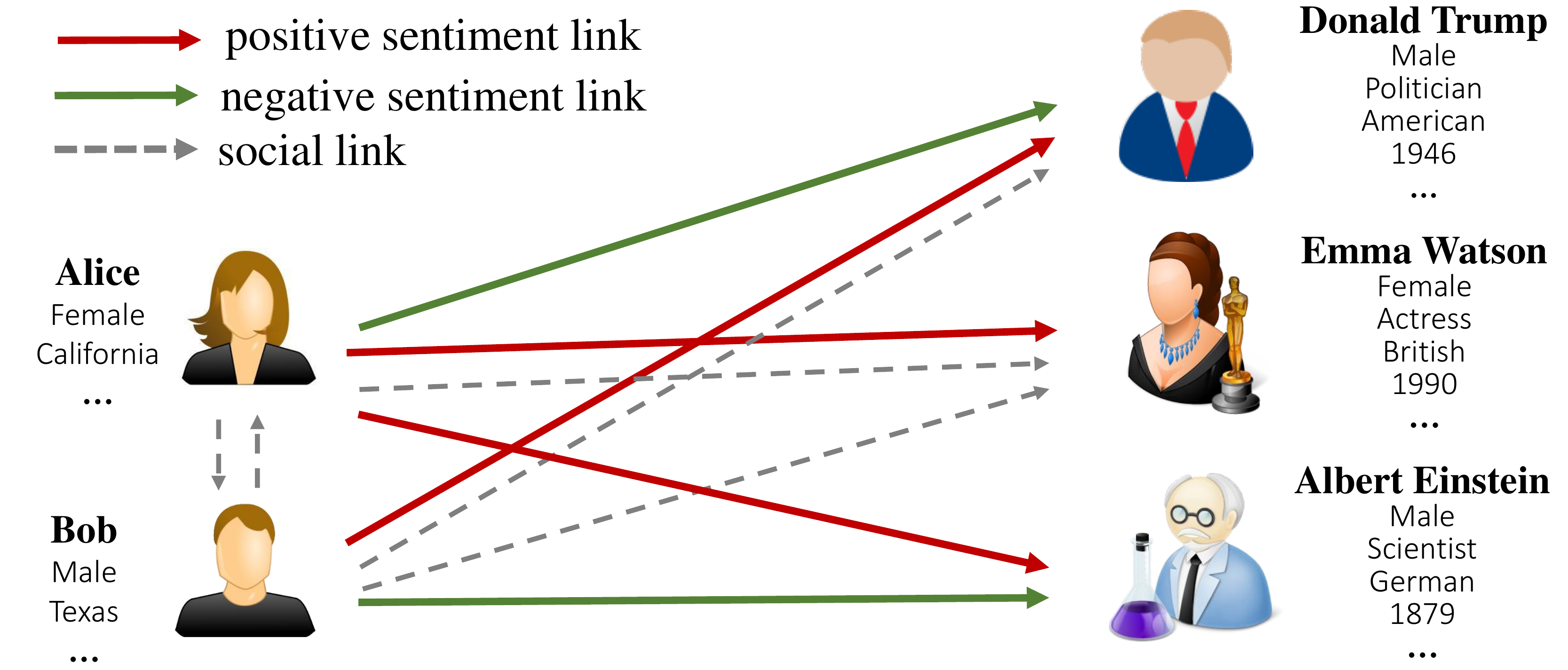}
  		\caption{Illustration of a snippet of heterogeneous networks with sentiment, social relationship and user profile.}
  		\label{fig:illustration}
	\end{figure}

	Based on the above facts, in this paper we investigate the problem of predicting sentiment links in absence of sentiment related content in online social networks.
	Our work is two-step.
	First, considering the lack of labeled data, we establish a labeled sentiment dataset from Weibo, one of the most popular social network sites in China.
	%The sentiment dataset contains 10$k$+ users and 100$k$+ tweets, where each tweet associates with a poster, a mentioned celebrity, and a binary label indicating the polarity of sentiment towards the celebrity.
	%We leverage state-of-the-art entity-level sentiment extraction method to calculate the polarity of sentiment.
	We leverage state-of-the-art entity-level sentiment extraction method to calculate the sentiment of the poster towards the celebrity in each tweet.
	Besides, to handle the sparsity problem, we collect two additional types of side information: social relationship among users and profile knowledge of users and celebrities.
	Our choices are enlightened by \cite{wang2017joint} and \cite{zhang2016collaborative}, respectively, in which \cite{wang2017joint} demonstrates that the structural information of social networks can greatly affect users' preference towards online items, and \cite{zhang2016collaborative} proves that information from knowledge base could boost the performance of recommendation.
	The heterogeneous information networks are illustrated in Fig. \ref{fig:illustration}.
	
	%The established dataset only provides us a small observed subset of the whole sentiment network.
	To explore more possible sentiment links from the network, in the second step, we propose a novel end-to-end framework termed as \textit{Signed Heterogeneous Information Network Embedding} (SHINE).
	Greatly different from existing network embedding approaches, SHINE is able to learn user representation and predict sentiment from signed heterogeneous networks.
	Specifically, SHINE adopts multiple deep autoencoders \cite{salakhutdinov2009semantic}, a type of deep-learning-based embedding technique, to extract users' highly nonlinear representations from the sentiment network, social network and profile network, respectively.
	The learned three types of user representations are subsequently fused together by specific aggregation function for further sentiment prediction.
	In addition to the adaptability to signed heterogeneous networks, the superiority of SHINE also lies in its end-to-end prediction technology and high flexibility of adding or removing modules of side information (i.e., social relationship and profile knowledge), which is discuss in Section \ref{section_shine}.
	
	We conduct extensive experiments on two real-world datasets.
	The results show that SHINE achieves substantial gains compared with baselines.
	Specifically, SHINE outperforms other strong baselines by $8.8\%$ to $16.8\%$ in the task of link prediction on Accuracy, and by $17.2\%$ to $219.4\%$ in the task of node recommendation on Recall@100 for positive nodes.
	The results also prove that SHINE is able to utilize the side information efficiently, and maintains a decent performance in cold start scenario.
	
%	In summary, the contributions of this paper are as follows:
%	\begin{itemize}
%		\item
%			We establish a labeled sentiment dataset with heterogeneous information from real-world online social networks by leveraging entity-level sentiment extraction methods.
%		\item
%			We propose SHINE framework, which extracts representation of users from sentiment network, social network and profile network, respectively, to predict the polarity of sentiment links among users.
%		\item
%			We conduct extensive experiments on two datasets, and the results reveal that our approach significantly outperforms state-of-the-art baseline methods.
%	\end{itemize}

\section{Related Work}\label{section_related_work}
%	\subsection{Sentiment Prediction in Social Networks}
%		Sentiment prediction in social networks are attracting researchers from various areas such as data mining, recommender systems, natural language processing, etc.
%		For example, \cite{das2014modeling, de2014learning, zhu2014tripartite, de2016learning} studies the problem of sentiment mining and forecasting in social networks.
%		\cite{zhang2014explicit, diao2014jointly, zhang2015incorporating} incorporate sentiment-level factors into recommender systems.
%		\cite{kiritchenko2014nrc, dos2014deep, nguyen2015phrasernn} propose different models to detect sentiment from web texts such tweets or product reviews.
%		Our work combines these ideas as we extract sentiment from web texts and do the prediction by leveraging structural information from social networks.

	\subsection{Signed Link Prediction}
		Our problem of predicting positive and negative sentiment links connects to a large body of work on signed social networks, including trust propagation \cite{guha2004propagation}, spectral analysis \cite{kunegis2010spectral}, and social media mining \cite{tang2016survey}.
		For the link prediction problem in signed graphs, Leskovec. \textit{et al.} \cite{leskovec2010predicting} adopt signed triads as features for prediction based on structural balance theory.
		Ye \textit{et al.} \cite{ye2013predicting} utilize transfer learning to leverage edge sign information from source network and improve prediction accuracy in target network.
		Tang \textit{et al.} design NeLP framework \cite{tang2015negative} which exploits positive links in social media to predict negative links.
		%Song \textit{et al.} \cite{song2015recommending} propose a new quantification method to improve recommendation performance for positive links.
		The difference between the above work and ours is that we construct a labeled dataset by entity-level sentiment extraction method, as there is no explicit signed links in mainstream online social networks.
		Besides, we use state-of-the-art deep learning approach to learn the representation of links.

	\subsection{Network Embedding}
		There is a long history of work on network embedding.
		Earlier works such as  IsoMap \cite{tenenbaum2000global} and Laplacian Eigenmap \cite{belkin2001laplacian} first construct the affinity graph of data using the feature vectors and then embed the affinity graph into a low-dimension space.
		Recently, DeepWalk \cite{perozzi2014deepwalk} deploys random walk to learning representations of social network.
		LINE \cite{tang2015line} proposes objective functions that preserve both local and global network structures for network embedding.
		Node2vec \cite{grover2016node2vec} designs a biased random walk procedure to learn a mapping of nodes that maximizes the likelihood of preserving network neighborhoods of nodes.
		SDNE \cite{wang2016structural} uses autoencoder to capture first-order and second-order network structures and learn user representation.
		However, these methods can only address unsigned and homogeneous networks.
		Additionally, several studies focus on representation learning in the scenario of heterogeneous network \cite{yu2014personalized, chang2015heterogeneous}, attributed network \cite{huang2017label}, or signed network \cite{wang2017signed, yuan2017sne}.
		However, these methods are specialized in only one particular type of networks, which is not applicable to the problem of sentiment prediction in real-world signed and heterogeneous networks.

\section{Dataset Establishment}\label{sec:data_collection_and_sentiment_extraction}
	In this section we introduce the process of collecting data from online social networks, and discuss the details of how to extract sentiment towards celebrities from tweets.
	
	\subsection{Data Collection}\label{sec:data_collection}
		\subsubsection{Weibo Tweets}
			We select Weibo\footnote{\url{http://weibo.com}} as the online social networks studied in this work.
			Weibo is one of the most popular social network sites in China which is akin to a hybrid of Facebook and Twitter.
			We collected 2.99 billion tweets on Weibo from August 14, 2009 to May 23, 2014 as raw dataset.
			To filter out useful data which contains sentiment towards celebrities, we first apply Jieba\footnote{\url{https://github.com/fxsjy/jieba}}, the most popular Chinese text segmentation tool, to tag the part of speech (POS) of each word for each tweet.
			%Then we establish a celebrity database (detailed in Section \ref{section_profile_of_celebrities}) and select tweets which contain words with POS tagging as ``person name''  appearing in the celebrity database.		
			Then we select those tweets containing words with POS tagging as ``person name'' which exist in our established celebrity database (detailed in Section \ref{sec:profile_of_celebrities}).
			After getting the set of candidate tweets, for each tweet we calculate its sentiment value (-1 to +1) towards the mentioned celebrities, and select those tweets with high absolute sentiment values.
			The final dataset consists of a set of triples $(a, b, s)$, where $a$ is the user who posts the tweet, $b$ is the certain celebrity mentioned in the tweet, and $s \in \{ +1, -1 \}$ is the sentiment polarity of user $a$ towards user $b$.
			The method of calculating sentiment values is detailed in Section \ref{sec:sentiment_extraction}.
			
		\subsubsection{Social Relation}
			In addition to the sentiment dataset, we also collect the social relation among users from Weibo.
			The dataset of social relation consists of tuples $(a, b)$, where $a$ is the follower and $b$ is the followee.
			
		\subsubsection{Profile of Ordinary Users}
			The profile of ordinary users are collected from Weibo.
			For each ordinary user, we extract two of his attributes, \textit{gender} and \textit{location}, as his profile information.
			The attribute values are represented as one-hot vectors.
			
		\subsubsection{Profile of Celebrities}\label{sec:profile_of_celebrities}
			We use Microsoft Satori\footnote{\url{http://searchengineland.com/library/bing/bing-satori}} knowledge base to extract profile of celebrities.
			First, we traverse the knowledge base and select terms with object type as ``person''.
			Then we filter out popular celebrities with high edit frequency in knowledge base and high appearance frequency in Weibo tweets.
			For each of these ``hot'' celebrities, we extract 9 attributes as his profile information: \textit{place of birth}, \textit{date of birth}, \textit{ethnicity}, \textit{nationality}, \textit{specialization}, \textit{gender}, \textit{height}, \textit{weight}, and \textit{astrological sign}.
			Values of these attributes are discretized so that every celebrity's attribute values can be expressed as one-hot vectors.
			Furthermore, we remove celebrities with ambiguous names as well as other noises.
	
			\begin{table}[t]
				\small
                		\centering
                		\caption{Statistics of Weibo sentiment datasets. ``celebrities v.'' means the celebrities owning verified accounts on Weibo.}
                		\begin{tabular}{|l|c||l|c|}
                    		\hline
                    	 	\# users & 12,814 & \# social links & 71,268 \\
                    		\hline
                    		\# celebrities & 1,723 & \# tweets & 126,380 \\
                    		\hline
                    		\# celebrities v.  & 706 & \# pos. tweets & 108,906 \\
                    		\hline
                    		\# ordinary users & 11,091 & \# neg. tweets & 17,474 \\
                    		\hline
				\end{tabular}
				\label{table:basic_statistics}
			\end{table}
		
	\subsection{Sentiment Extraction}\label{sec:sentiment_extraction}
		To extract users' sentiment towards celebrities in tweets, we first generate a sentiment lexicon consisting of words and their \textit{sentiment orientation} (SO) scores.
		To achieve this, we manually construct a emoticon-sentiment mapping file and map each tweet to positive or negative class according to the label of emoticon appeared in the tweet.
		For example, \textit{``I love Kobe! [kiss]''} is mapped to positive class if the key-value pair ([kiss], \textit{positive}) exists in the emoticon-sentiment mapping file.
		Note that the class of emoticon cannot be directly regarded as the sentiment towards celebrities since we found a large number of mismatch cases, e.g., \textit{``Miss you Taylor Swift [cry][cry]''}.
		Afterwards, for each word (segmented by Jieba) with occurrence frequency from 2,000 to 10,000,000 in the raw tweets datasets, similar to \cite{bravo2015positive}, we calculate its SO score as
		\begin{equation}
			SO(word) = PMI(word, pos) - PMI(word, neg),
		\end{equation}
		where PMI is the \textit{point-wise mutual information} \cite{turney2002thumbs} defined as $PMI(x, y) = \log \frac{p(x, y)}{p(x) p(y)}$, $pos$ and $neg$ are the tweets of positive and negative class, respectively.
		SO scores are subsequently normalized to $[-1, 1]$.
		
		After getting the lexicon, we use SentiCircle \cite{saif2015semantic} to calculate sentiment towards celebrities in each tweet.
		Given a piece of tweet as well as the mentioned celebrity, we represent the contextual semantics of the celebrity as a polar coordinate space, where the celebrity is situated in the origin and other terms in the tweet are scattered around.
		Specifically, for celebrity term $c$, the coordinate of term $t_i$ is $(r_i, \theta_i)$, where $r_i$ is the inverse of distance between $c$ and $t_i$ in syntax dependence graph generated by LTP \cite{che2010ltp}, and $\theta_i = SO(t_i) \cdot \pi$.
		The overall sentiment towards the celebrity $c$ is, therefore, approximated as the geometric center of all terms $c_i$.
		We take the projection of the geometric center on y-axis as final sentiment value towards the celebrity.
		
		To validate the effectiveness of sentiment extraction, we randomly select 1,000 tweets (500 positive and 500 negative tagged by our method) in Weibo sentiment dataset, and manually label each one of them.
		The result shows that the precision is $95.2\%$ for positive class and $91.0\%$ for negative class, which we believe is accurate enough for subsequent experiments.
		The basic statistics of Weibo sentiment datasets is presented in Table \ref{table:basic_statistics}.

	\begin{figure}[t]
		\centering
		\begin{subfigure}{.15\textwidth}
			\centering
			\includegraphics[width=\textwidth]{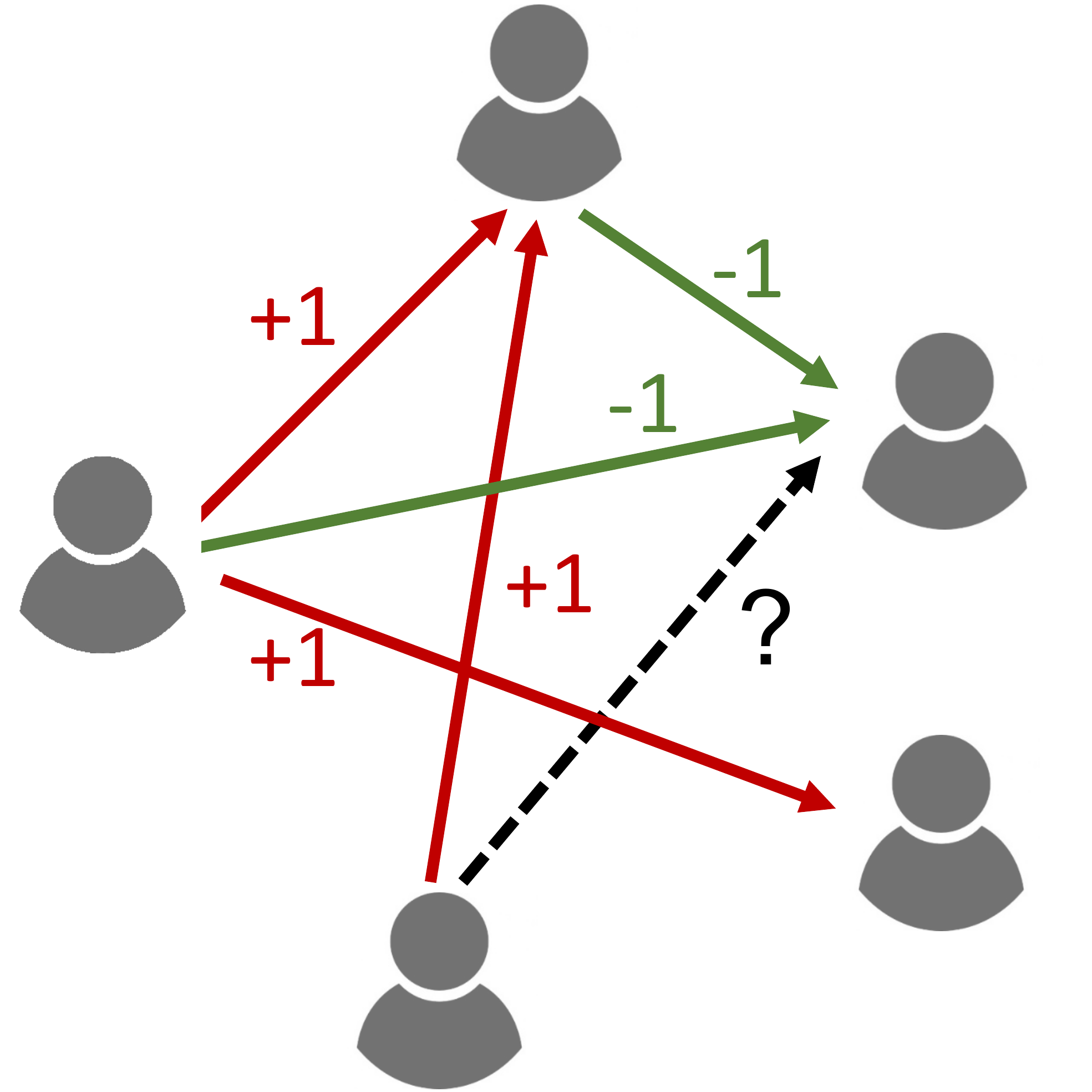}
			\caption{Sentiment network}
		\end{subfigure}
		\hfill
		\begin{subfigure}{.16\textwidth}
			\centering
			\includegraphics[width=0.94\textwidth]{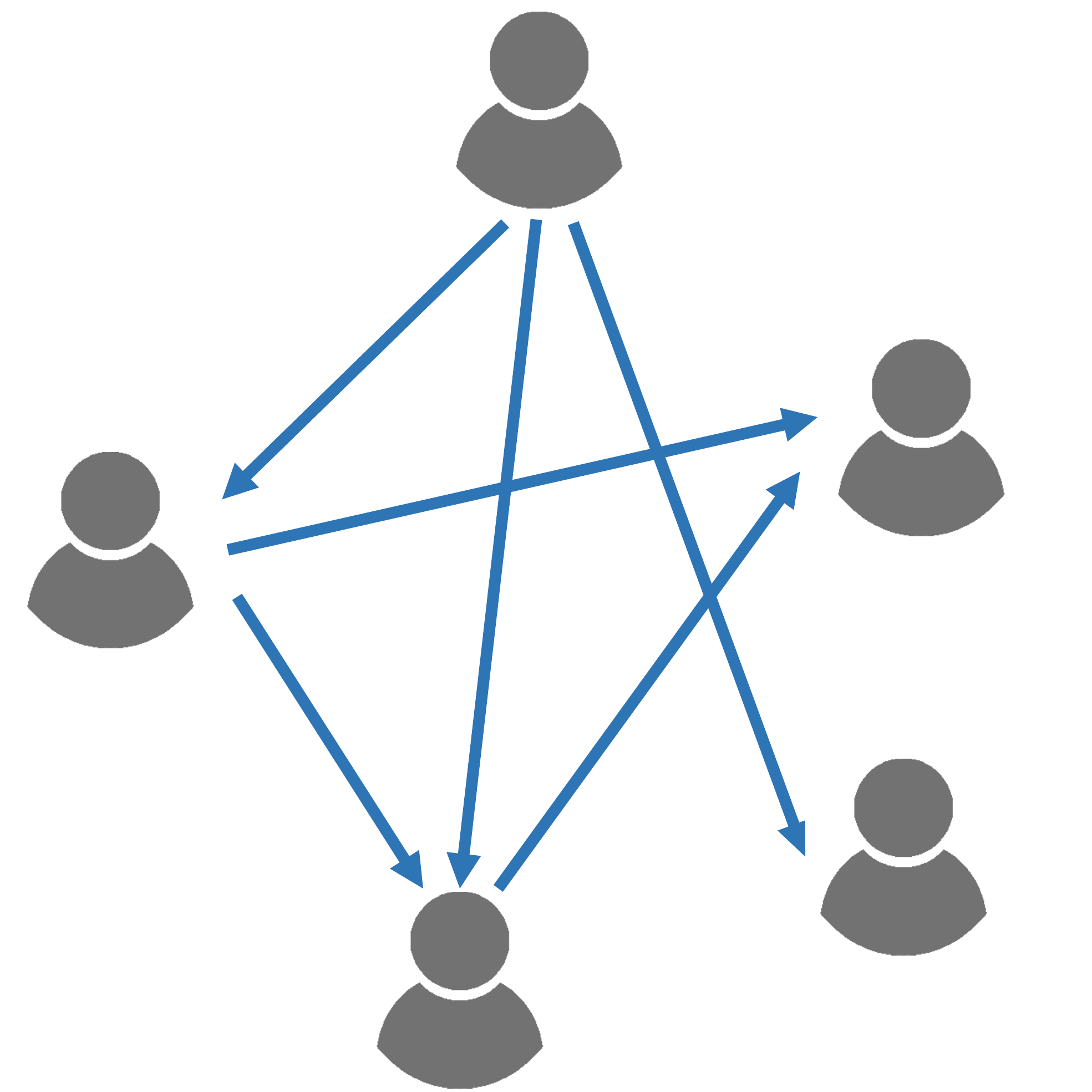}
			\caption{Social network}
		\end{subfigure}
		\hfill
		\begin{subfigure}{.15\textwidth}
			\centering
			\includegraphics[width=0.91\textwidth]{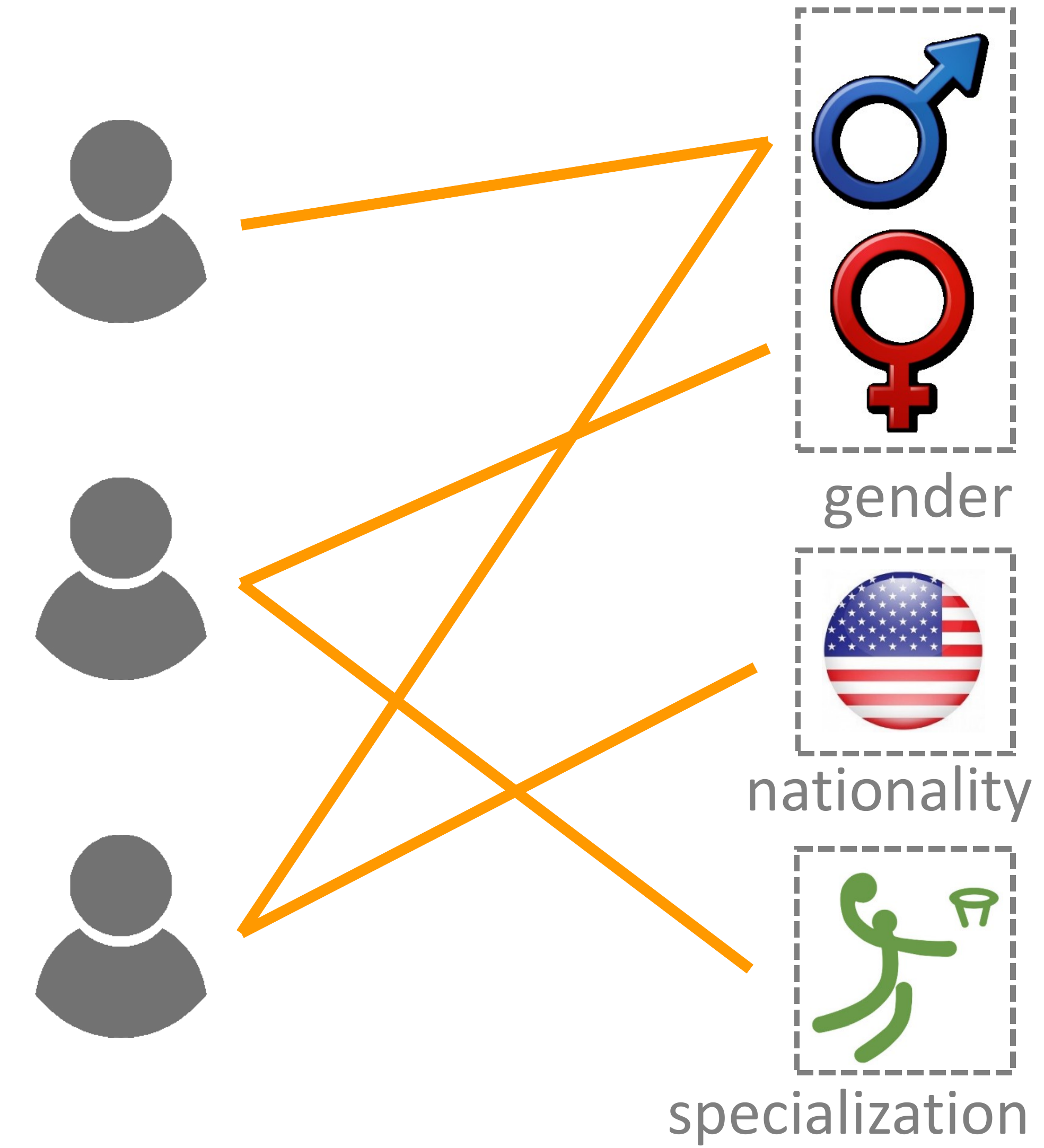}
			\caption{Profile network}
		\end{subfigure}
  		\caption{Illustration of the three studied networks.}
  		\label{fig:three_networks}
	\end{figure}

	\begin{figure*}[t]
		\centering
  		\includegraphics[width=.95\textwidth]{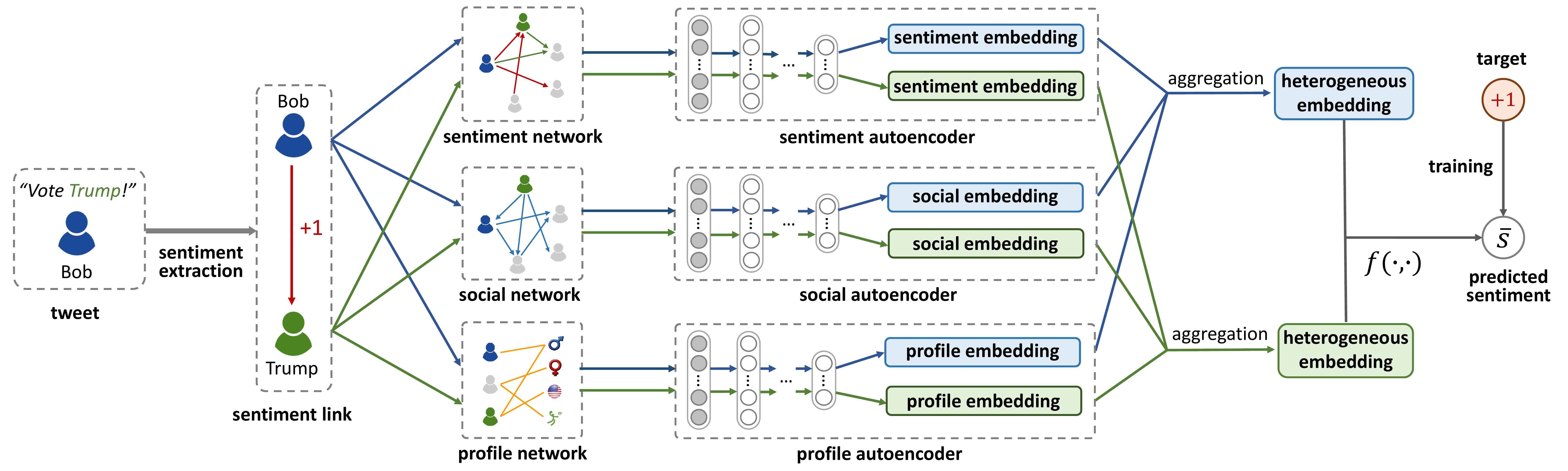}
  		\caption{Framework of the end-to-end SHINE model. To clearly demonstrate the model, we only show the encoder part of all the three autoencoders and leave out the decoder part in this figure.}
  		\label{fig:framework}
	\end{figure*}

\section{Problem Formulation}\label{sec:problem_formulation}
	\label{problem_formulation}
	In this section we formulate the problem of predicting sentiment links in heterogeneous information networks.
	For better illustration, we split the original heterogeneous network into the following three single-type networks:
	
	\textbf{Sentiment network.}
	The directed \textit{sentiment network} is denoted as $G_s = (V, S)$, where $V = \{ 1, ..., |V| \}$ represents the set of users (either ordinary users or celebrities) and $S = \{s_{ij} \ | \ i \in V, j \in V \}$ represents sentiment links among users.
	Each $s_{ij}$ can take the value of $+1$, $-1$ or $0$, representing that user $i$ holds a positive, negative, or unobserved sentiment towards user $j$, respectively.
	
	\textbf{Social network.}
	The directed \textit{social network} is denoted as $G_r = (V, R)$, where $R = \{r_{ij} \ | \ i \in V, j \in V \}$ represents social links among users.
	Each $r_{ij}$ can take the value of 1 or 0, representing that user $i$ follows user $j$ or not in the social network.
	
	\textbf{Profile network.}
	We denote $\mathcal A = \{ A_1, ..., A_{|\mathcal A|} \}$ the set of user's attributes, and 
	$a_{kl} \in A_k$ the $l$-th possible value of attribute $A_k$.
	We take the union of all possible values of attributes and renumber them as $U = \bigcup A_k = \{ a_j \ | \ j = 1, ..., \sum_k |A_k| \}$.
	Then the undirected bipartite \textit{profile network} can be denoted as $G_p = (V, U, P)$, where $P = \{p_{ij} \ | \ i \in V, a_j \in U \}$ represents profile links between users and attribute values.
	Each $p_{ij}$ can take the value of 1 or 0, representing that user $i$ possesses attribute value $j$ or not.
	
	The three networks are illustrated in Fig. \ref{fig:three_networks}.

	\textbf{Sentiment links prediction.}
	We define the problem of predicting sentiment links in heterogeneous information networks as follows:
	Given the sentiment network $G_s$, social network $G_r$ and profile network $G_p$, we aim to predict the sentiment of unobserved links between users in $G_s$.

\section{Signed Heterogeneous Information Network Embedding}\label{section_shine}
	In this section we introduce the proposed SHINE model.
	We first show the whole framework of SHINE.
	Then we present the details of the SHINE model, including how to extract user representation jointly from the three networks as well as the learning algorithm.
	At last we give some discussions on the model.

	\subsection{Framework}
		In this paper we propose an end-to-end SHINE model to predict sentiment links.
		The framework of SHINE is shown in Fig. \ref{fig:framework}.
		In general, the whole framework consists of three major components: sentiment extraction and heterogeneous networks construction (the left part), user representation extraction (the middle part), as well as representation aggregation and sentiment prediction (the right part).
		For each tweet mentioning a specific celebrity, we first calculate the associated sentiment (discussed in Section \ref{sec:data_collection}), and represent the user and the celebrity in this sentiment link by using their neighborhood information from the three constructed networks (introduced in Section \ref{sec:problem_formulation}).
		We then design three distinct autoencoders to extract short and dense embeddings from original sparse neighborhood-based representation respectively, and aggregate these three kinds of embeddings into final heterogeneous embedding.
		The predicted sentiment can thus be calculated by applying specific similarity measurement function (e.g., inner product or logistic regression) to the two heterogeneous embeddings, and the whole model can be trained based on the predicted sentiment and the target (i.e., the ground truth obtained in sentiment extraction step).
		In the following subsections we will introduce SHINE model in detail.

	\subsection{Sentiment Network Embedding}\label{sec:sentiment_embedding}
		Given the sentiment graph $G_s = (V, S)$, for each user $i \in V$, we define its sentiment adjacency vector ${\bf x}_i = \{ s_{ij} \ | \ j \in V \} \cup \{ s_{ji} \ | \ j \in V \}$.
		Note that ${\bf x}_i$ fully contains the global incoming and outgoing sentiment information of user $i$.
		However, it is impractical to take ${\bf x}_i$ directly as the sentiment representation of user $i$, as the adjacency vector is too long and sparse for further processing.
		Recently, a lot of network embedding models \cite{perozzi2014deepwalk, tang2015line, grover2016node2vec, wang2016structural} are proposed, which aim to learn low-dimension representations of vertices while preserving the network structure.
		Among those models, deep autoencoder is proved to be one of state-of-the-art solutions, as it is able to capture highly nonlinear network structure by using deep models \cite{wang2016structural}.
		In general, autoencoder \cite{salakhutdinov2009semantic} is an unsupervised neural network model of codings aiming to learn a representation of a set of data.
		Autoencoder consists of two parts, the encoder and the decoder, which contains multiple nonlinear functions (layers) for mapping the input data to representation space and reconstructing original input from representation, respectively.
		In our SHINE model, we propose to use autoencoders for efficiently user representation learning.
		
\begin{figure}[t]
		\centering
  		\includegraphics[width=.48\textwidth]{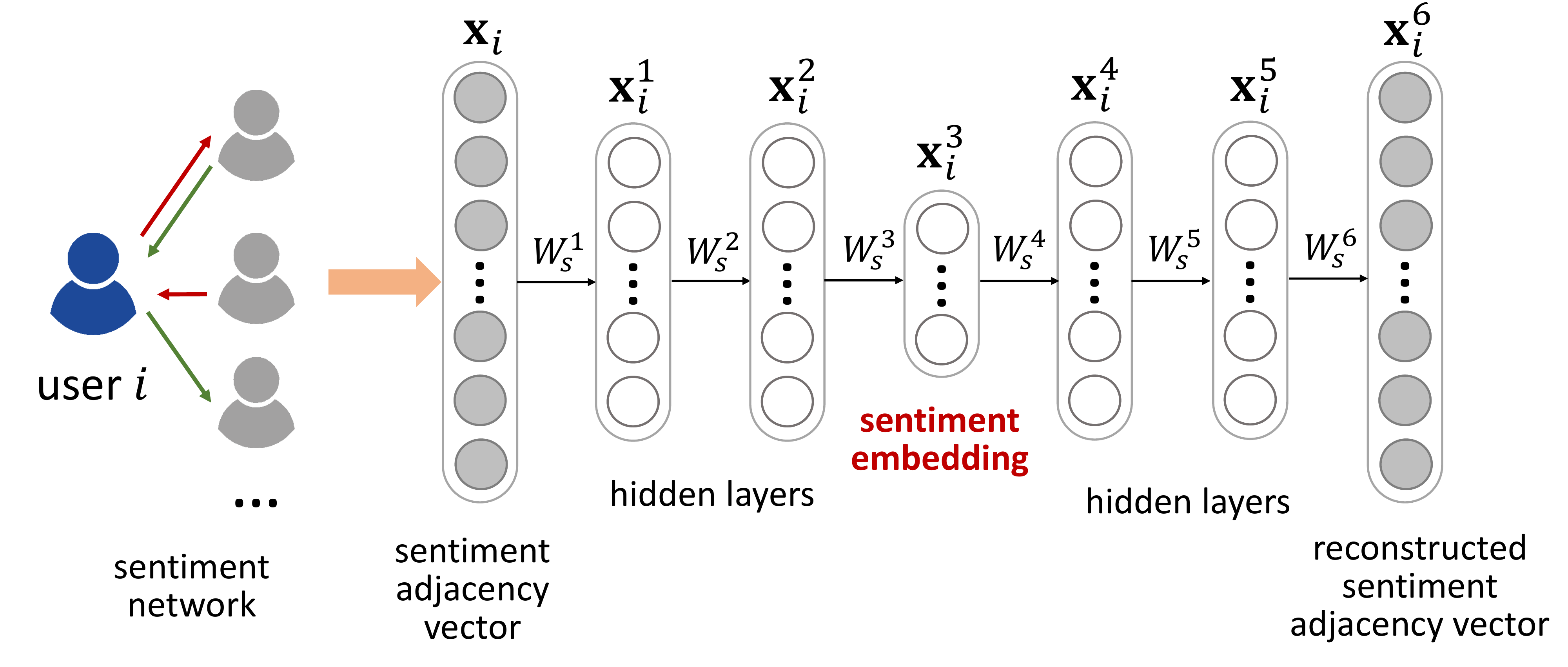}
  		\caption{Illustration of a 6-layer autoencoder for sentiment network embedding.}
  		\label{fig:sentiment_graph_embedding}
	\end{figure}

		Fig. \ref{fig:sentiment_graph_embedding} illustrates the autoencoder for sentiment network embedding.
		As shown in Fig. \ref{fig:sentiment_graph_embedding}, the sentiment autoencoder maps each user to a low-dimension latent representation space and recover original information from latent representation by using multiple fully-connected layers.
		Given the input ${\bf x}_i$, the hidden representations for each layer are
		\begin{equation}\label{autoencoder_sentiment}
			{\bf x}_i^k = \ \sigma \left( {\bf W}_s^k {\bf x}_i^{k-1} + {\bf b}_s^k \right), \ k = 1, 2, ..., K_s,
		\end{equation}
		where ${\bf W}_s^k$ and ${\bf b}_s^k$ are weight and bias parameters of layer $k$ in the sentiment autoencoder, respectively, $\sigma(\cdot)$ is the nonlinear activation function, $K_s$ is the number of layers of sentiment autoencoder, and ${\bf x}_i^0 = {\bf x}_i$.
		For simplicity, we denote ${\bf x}_i' = {\bf x}_i^{K_s}$ the reconstruction of ${\bf x}_i$.
		
		The basic goal of the autoencoder is to minimize the reconstruction loss between input and output representations.
		Similar to \cite{wang2016structural}, in SHINE model the reconstruction loss term of sentiment autoencoder is defined as
		\begin{equation}
			\mathcal L_s = \sum\nolimits_{i \in V} \left\| ({\bf x}_i - {\bf x}_i') \odot {\bf l}_i \right\|_2^2,
		\end{equation}
		where $\odot$ denotes the Hadamard product, and ${\bf l}_i = (l_{i,1}, l_{i,2}, ..., l_{i,2|V|})$ is the sentiment reconstruction weight vector in which
		\begin{equation}
			l_{i,j} =
			\begin{cases}
				\alpha > 1, & if \ s_{ij} = \pm 1; \\
				1, & if \ s_{ij} = 0.
			\end{cases}
		\end{equation}
		
		The meaning of the above loss term lies in that we impose more penalty to the reconstruction error of the non-zero elements than that of zero elements in input $\bf{x}_i$, as a non-zero $s_{ij}$ carries more explicit sentiment information than an implicit zero $s_{ij}$.
		Note that the sentiment embedding of user $i$ can be obtained from the layer $K_s / 2$ in the sentiment autoencoder, and we denote ${\bf\widehat x}_i = {\bf x}_i^{K_s / 2}$ the sentiment embedding of user $i$ for simplicity.

	\subsection{Social Network Embedding}
		Similar to previous sentiment network embedding, we apply autoencoder to extract user representation from the social network.
		Given the social network $G_r = (V, R)$, for each user $i \in V$, we define its social adjacency vector ${\bf y}_i = \{ r_{ij} \ | \ j \in V \} \cup \{ r_{ji} \ | \ j \in V \}$, which fully contains the structural information of user $i$ in the social network.
		The hidden representations of each layer in the social autoencoder are
		\begin{equation}
			{\bf y}_i^k = \ \sigma \left( {\bf W}_r^k {\bf y}_i^{k-1} + {\bf b}_r^k \right), \ k = 1, 2, ..., K_r,
		\end{equation}
		where the meaning of notations are similar to those in Eq. (\ref{autoencoder_sentiment}).
		We also denote ${\bf y}_i' = {\bf y}_i^{K_r}$ the reconstruction of ${\bf y}_i$.
		Similarly, the reconstruction loss term of social autoencoder is
		\begin{equation}
			\mathcal L_r = \sum\nolimits_{i \in V} \left\| ({\bf y}_i - {\bf y}_i') \odot {\bf m}_i \right\|_2^2,
		\end{equation}
		where ${\bf m}_i = (m_{i,1}, m_{i,2}, ..., m_{i,2|V|})$ is the social reconstruction weight vector in which if $r_{ij} = 1$, $m_{i,j} = \alpha > 1$, else $m_{i,j} = 1$.
		The social embedding of user $i$ is denoted as ${\bf\widehat y}_i = {\bf y}_i^{K_r / 2}$.
	
	\subsection{Profile Network Embedding}	
		The profile network $G_p = (V, U, P)$ is an undirected bipartite graph which consists of two disjoint sets of users and attribute values.
		For each user $i \in V$, its profile adjacency vector is defined as ${\bf z}_i = \{ p_{ij} \ | \ j \in U \}$.
		User $i$'s hidden representations of each layer in the profile autoencoder are
		\begin{equation}
			{\bf z}_i^k = \ \sigma \left( {\bf W}_{p}^k {\bf z}_i^{k-1} + {\bf b}_{p}^k \right), \ k = 1, 2, ..., K_{p},
		\end{equation}
		where the meaning of notations are similar to those in Eq. (\ref{autoencoder_sentiment}).
		We also use the notation ${\bf z}_i'$ to denote the reconstruction of ${\bf z}_i$.
		Therefore, the reconstruction loss term of profile autoencoder is
		\begin{equation}
			\mathcal L_p = \sum\nolimits_{i \in V} \left\| ({\bf z}_i - {\bf z}_i') \odot {\bf n}_i \right\|_2^2,
		\end{equation}
		where ${\bf n}_i$ is the profile reconstruction weight vector defined similarly to ${\bf m}_i$ in the previous subsection.
		The profile embedding of user $i$ is denoted as ${\bf\widehat z}_i = {\bf z}_i^{K_p / 2}$.

	\subsection{Representation Aggregation and Sentiment Prediction}\label{sec:sentiment_prediction}
		Once we obtain the sentiment embedding ${\bf\widehat x}_i$, social embedding ${\bf\widehat y}_i$, and profile embedding ${\bf\widehat z}_i$ of user $i$, we can aggregate these embeddings into final heterogeneous embedding ${\bf e}_i$ by specific aggregation function $g(\cdot, \cdot, \cdot)$.
		We list some of the available aggregation functions as follows:
		\begin{itemize}
		\item
			\textit{Summation} \cite{zhang2016collaborative}, i.e., ${\bf e}_i = {\bf\widehat x}_i + {\bf\widehat y}_i + {\bf\widehat z}_i$;
		\item
			\textit{Max pooling} \cite{wang2015learning}, i.e., ${\bf e}_i = element\textrm{-}wise\textrm{-}max({\bf\widehat x}_i, {\bf\widehat y}_i, {\bf\widehat z}_i)$;
		\item
			\textit{Concatenation} \cite{tang2015line}, i.e., ${\bf e}_i = \langle {\bf\widehat x}_i, {\bf\widehat y}_i, {\bf\widehat z}_i \rangle$.
	\end{itemize}
	
		Finally, given two users $i$ and $j$ as well as their heterogeneous embedding ${\bf e}_i$ and ${\bf e}_j$, the predicted sentiment $\bar s_{ij}$ can be calculated as $\bar s_{ij} = f(i, j)$, where $f(\cdot, \cdot)$ is specific similarity measurement function.
		For example:
		\begin{itemize}
		\item
			\textit{Inner product} \cite{chang2015heterogeneous, dong2017hybrid}, i.e., $\bar s_{ij} = {\bf e}_i^\textrm{T} {\bf e}_j + b$, where $b$ is a trainable bias parameter;
		\item
			\textit{Euclidean distance} \cite{wang2016structural}, i.e., $\bar s_{ij} = -\| {\bf e}_i - {\bf e}_j \|_2 + b$, where $b$ is a trainable bias parameter;
		\item
			\textit{Logistic regression} \cite{perozzi2014deepwalk}, i.e., $\bar s_{ij} = {\bf W}^\textrm{T} \langle {\bf e}_i, {\bf e}_j \rangle + b$, where $\bf W$ and $b$ are trainable weights and bias parameters.
	\end{itemize}
	
	We will study the choices of $f$ and $g$ in the experimental part.

	\subsection{Optimization}
		The complete objective function of SHINE model is as follows:
		\begin{equation}
		\label{eq:objective_function}
			\begin{split}
				\mathcal L =& \sum\nolimits_{i \in V} \left\| ({\bf x}_i - {\bf x}_i') \odot {\bf l}_i \right\|_2^2 + \lambda_1 \sum\nolimits_{i \in V} \left\| ({\bf y}_i - {\bf y}_i') \odot {\bf m}_i \right\|_2^2\\
				& + \lambda_2 \sum\nolimits_{i \in V} \left\| ({\bf z}_i - {\bf z}_i') \odot {\bf n}_i \right\|_2^2 + \lambda_3 \sum\nolimits_{s_{ij} = \pm 1} \left( f({\bf e}_i, {\bf e}_j) - s_{ij} \right)^2\\
				& + \lambda_4 \mathcal L_{reg},
			\end{split}
		\end{equation}
		where $\lambda_1$, $\lambda_2$, $\lambda_3$ and $\lambda_4$ are balancing parameters.
		The first three terms in Eq. (\ref{eq:objective_function}) are the reconstruction loss terms of sentiment autoencoder, social autoencoder, and profile autoencoder, respectively.
		The fourth term in Eq. (\ref{eq:objective_function}) is the supervised loss term for penalizing the divergence between predicted sentiment and ground truth.
		The last term in Eq. (\ref{eq:objective_function}) is the regularization term that prevents over-fitting, i.e.,
		\begin{equation}
		\label{eq:regularization}
			\begin{split}
				\mathcal L_{reg} = \sum_{k = 1}^{K_s} \big\| {\bf W}_s^k \big\|_2^2 + \sum_{k = 1}^{K_r} \big\| {\bf W}_r^k \big\|_2^2 + \sum_{k = 1}^{K_p} \big\| {\bf W}_p^k \big\|_2^2 + \big\| f \big\|^2_2,
			\end{split}
		\end{equation}
		where ${\bf W}_s^k$, ${\bf W}_r^k$, ${\bf W}_p^k$ are the weight parameters of layer $k$ in the sentiment autoencoder, social autoencoder, and profile autoencoder, respectively, and $\| f \|^2_2$ is the regularization penalty for similarity measurement function $f(\cdot, \cdot)$ (if appropriate).
		
		We employ the AdaGrad \cite{duchi2011adaptive} algorithm to minimize the objective functions in Eq. (\ref{eq:objective_function}).
		In each iteration, we randomly select a batch of sentiment links from training dataset and compute the gradient of the objective function with respect to each trainable parameter respectively.
		Then we update each trainable parameter according to the AdaGrad algorithm till convergence.

	\subsection{Discussions}
		\subsubsection{Asymmetry}
		Many real-world networks are directed, which implies that for two nodes $i$ and $j$ in the network, edges $(i, j)$ and $(j, i)$ may coexist and their values are not necessarily identical.
		A few recent studies have focused on this asymmetry issue \cite{ou2016asymmetric, zhou2017scalable}.
		In this work, whether the basic SHINE model can characterize asymmetry depends on the choice of similarity measurement function $f$.
		Specifically, SHINE is capable of dealing with the direction of a link if and only if $f(i,j) \neq f(j,i)$ (e.g., logistic regression).
		However (and fortunately), even if we choose a symmetric function (e.g., inner product or Euclidean distance) as $f$, we can still easily extend the basic SHINE model to asymmetry-aware version by setting two distinct sets of autoencoders to extract representation of source node and target node respectively.
		From this point of view, in basic SHINE model the parameters of autoencoders are actually \textit{shared} for source node and target node to alleviate over-fitting, and we can choose to explicitly distinguish the two sets of autoencoders for asymmetry reasons.
		
		\subsubsection{Cold start problem}
			A practical issue for network embedding is how to learn representations for newly arrived node, which is the cold start problem.
			Almost all existing models cannot work well in cold start scenario because they only use the information from the target network (e.g., sentiment network in this paper), which is not applicable for the newly arrived node who has little interaction with the existing target network.
			However, SHINE is free of the cold start problem, as it makes full use of side information and incorporate it naturally into the target network when learning user representations.
			We will further study the performance of SHINE in cold start scenario in the experiment part.
		
		\subsubsection{Flexibility}
			It is worth noticing that SHINE is also a framework with high flexibility.
			For any other new available side information of users (e.g., users' browsing history), we can easily design a new parallel processing component and ``plug'' it in the original SHINE framework to assist learning representation.
			Contrarily, we can also ``pull out'' social autoencoder or profile autoencoder from SHINE framework if such side information is unavailable.
			Besides, the flexibility of SHINE also lies in that one can choose different aggregation functions $g$ and similarity measurement functions $f$, as discussed in Section \ref{sec:sentiment_prediction}.

\section{Experiments}\label{section_experiments}
	In this section, we evaluate the performance of our proposed SHINE on real-world datasets.
	We first introduce the datasets, baselines, and parameter settings for experiments, then present the experimental results of SHINE and baselines.	
	
	\subsection{Datasets}
		To comprehensively demonstrate the effectiveness of SHINE framework, we use the following two datasets for experiments:
		\begin{itemize}
			\item
				\textbf{Weibo-STC}: Our proposed \textit{Weibo Sentiment Towards Celebrities} dataset consists of three heterogeneous networks with 12,814 users, 126,380 tweets, 71,268 social links and 37,689 profile values, of which the detail is presented in Section \ref{sec:data_collection_and_sentiment_extraction}.
			\item
				\textbf{Wiki-RfA}: \textit{Wikipedia Requests for Adminship} \cite{west2014exploiting} is a signed network with 10,835 nodes and 159,388 edges, corresponding to votes cast by Wikipedia uses in election for promoting individuals to the role of administrator.
				A signed link indicates a positive or negative vote by one user on the promotion of another.
				Note that Wiki-RfA does not contain any side information of nodes, therefore, this dataset is used to validate the efficacy of the basic sentiment autoencoder in SHINE.
		\end{itemize}
	
%	\subsection{Evaluation Metrics}
%		To quantitatively analyze the performance of link prediction, similar to \cite{zhang2016collaborative}, in our experiment we use \textit{Mean Average Precision} (\textit{MAP}) and \textit{Recall} as the evaluation metrics.
%		Since the prediction results include positive and negative class, we choose $pMAP@K$, $nMAP@K$, $pRecall@K$, and $nRecall@K$ to evaluate the performance on top $k$ prediction in positive and negative cases, where ``\textit{p}'' denotes \textit{positive} and ``\textit{n}'' denotes \textit{negative}.

		\subsection{Baselines}
		We use the following five methods as baselines, in which the first three are network embedding methods, FxG is a signed link prediction approach, and LIBFM is a generic classification model.
		Note that the first three methods are not directly applicable to signed heterogeneous networks, so we use them to learn user representations from positive and negative part of each network respectively, and concatenate them to form the final embeddings.
		For FxG on Weibo-STC dataset, we only use the sentiment network as input because the FxG model cannot utilize the side information of nodes.
		\begin{itemize}
			\item
				\textbf{LINE}: Large-scale Information Network Embedding \cite{tang2015line} defines loss functions to preserve the first-order and second-order proximity and learns representations of vertices.
			\item
				\textbf{Node2vec}: Node2vec \cite{grover2016node2vec} designs a biased random walk procedure to learn a mapping of nodes that maximizes the likelihood of preserving network neighborhoods of nodes.
			\item
				\textbf{SDNE}: Structural Deep Network Embedding \cite{wang2016structural} is a semi-supervised network embedding model using autoencoder to capture local and global structure of target networks.
			\item
				\textbf{FxG}: Fairness and Goodness \cite{kumar2016edge} predicts the weights of edges in weighted signed networks by introducing two measures of node behavior: goodness (i.e., how much the node is liked by other nodes) and fairness (i.e., how fair the node is in rating other nodes' likeability).
			\item
				\textbf{LIBFM}: LIBFM \cite{rendle2012factorization}  is a state-of-the-art feature based factorization model.
				In this paper, we use the concatenated one-hot vectors of users in three networks as input to feed LIBFM.
		\end{itemize}

		\begin{figure*}[t]
			\centering
			\begin{subfigure}{.9\textwidth}
				\vspace*{-0.1in}
				\centering
				\includegraphics[width=\textwidth]{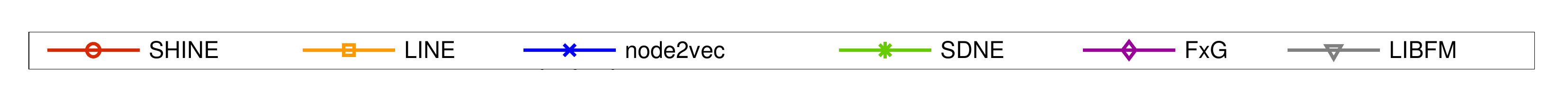}
				\vspace*{-0.25in}
			\end{subfigure}
			\hfill
			\begin{subfigure}{.24\textwidth}
				\centering
				\includegraphics[width=\textwidth]{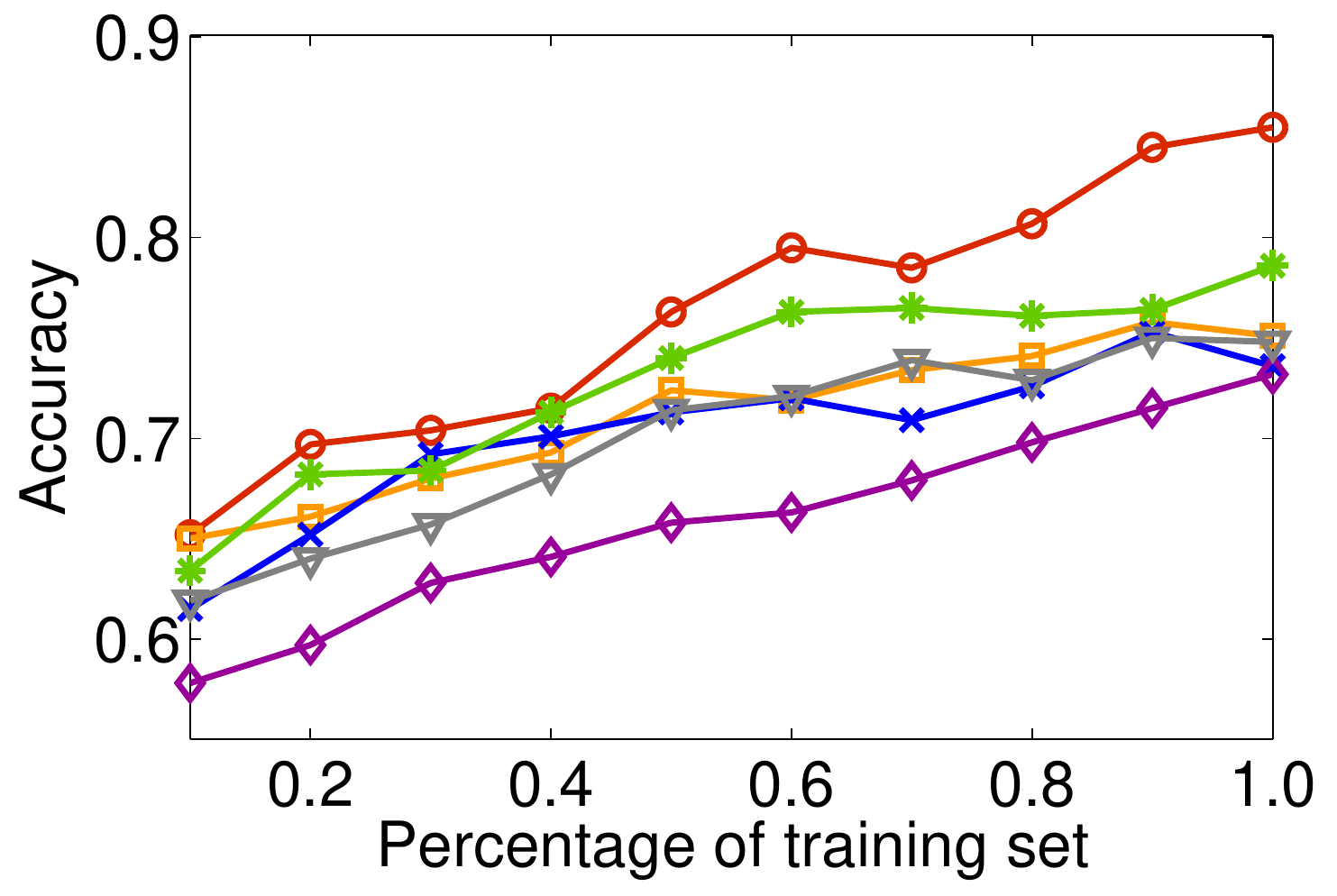}
				\vspace*{-0.1in}
				\caption{Accuracy on Weibo-STC}
			\end{subfigure}
			\hfill
			\begin{subfigure}{.24\textwidth}
				\centering
				\includegraphics[width=\textwidth]{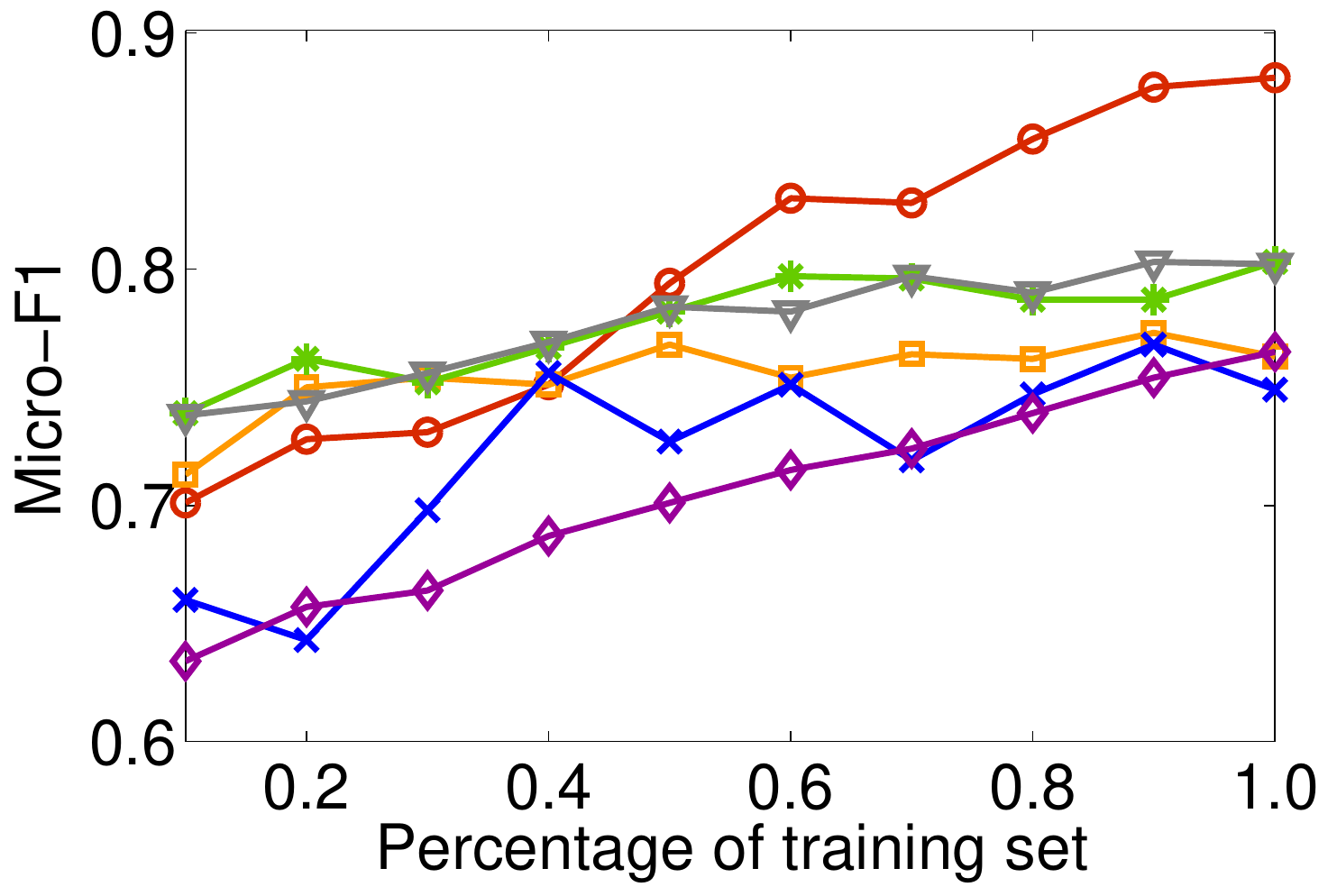}
				\vspace*{-0.1in}
				\caption{Micro-F1 on Weibo-STC}
			\end{subfigure}
			\hfill
			\begin{subfigure}{.24\textwidth}
				\centering
				\includegraphics[width=\textwidth]{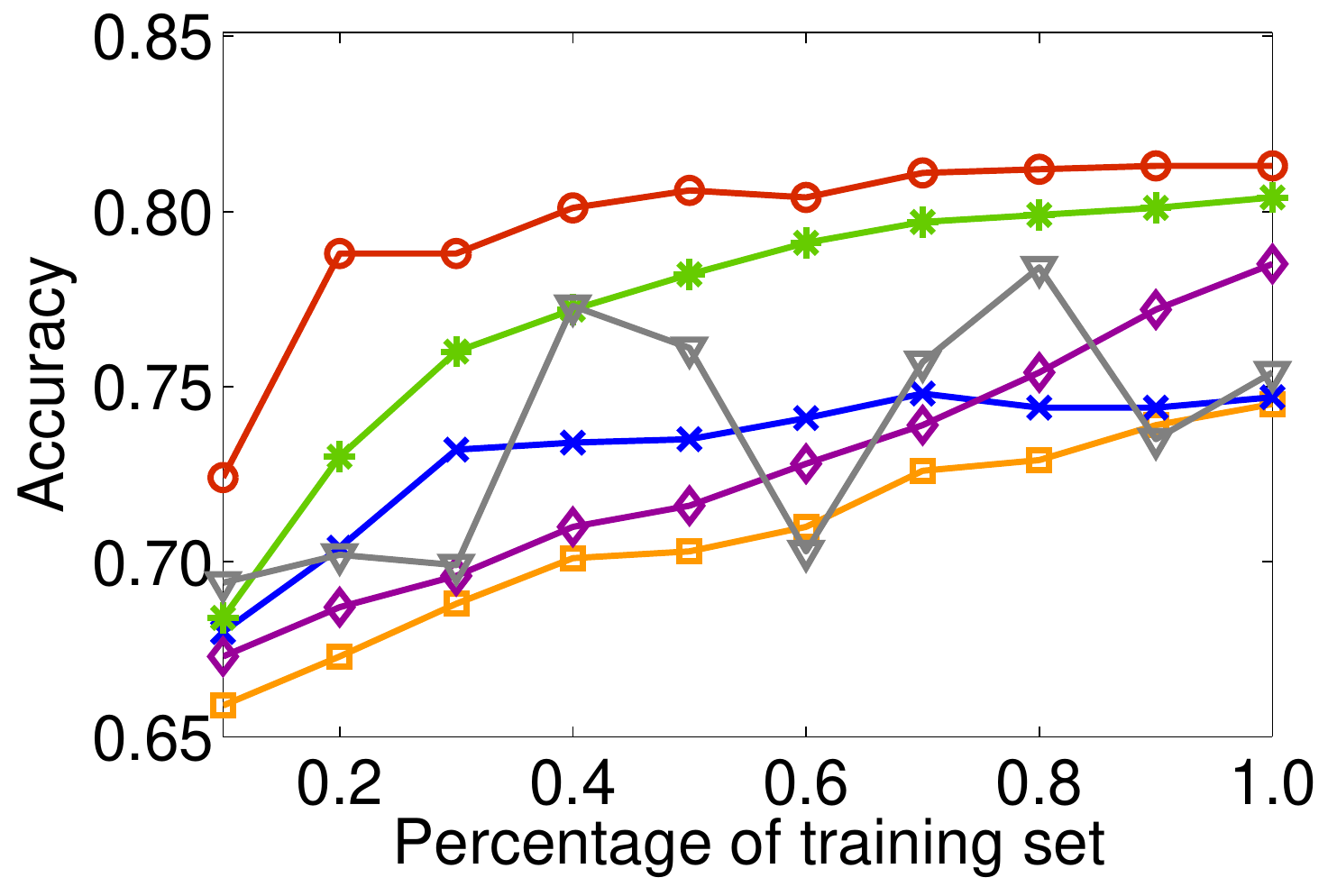}
				\vspace*{-0.1in}
				\caption{Accuracy on Wiki-RfA}
				\label{fig:5c}
			\end{subfigure}
			\hfill
			\begin{subfigure}{.24\textwidth}
				\centering
				\includegraphics[width=\textwidth]{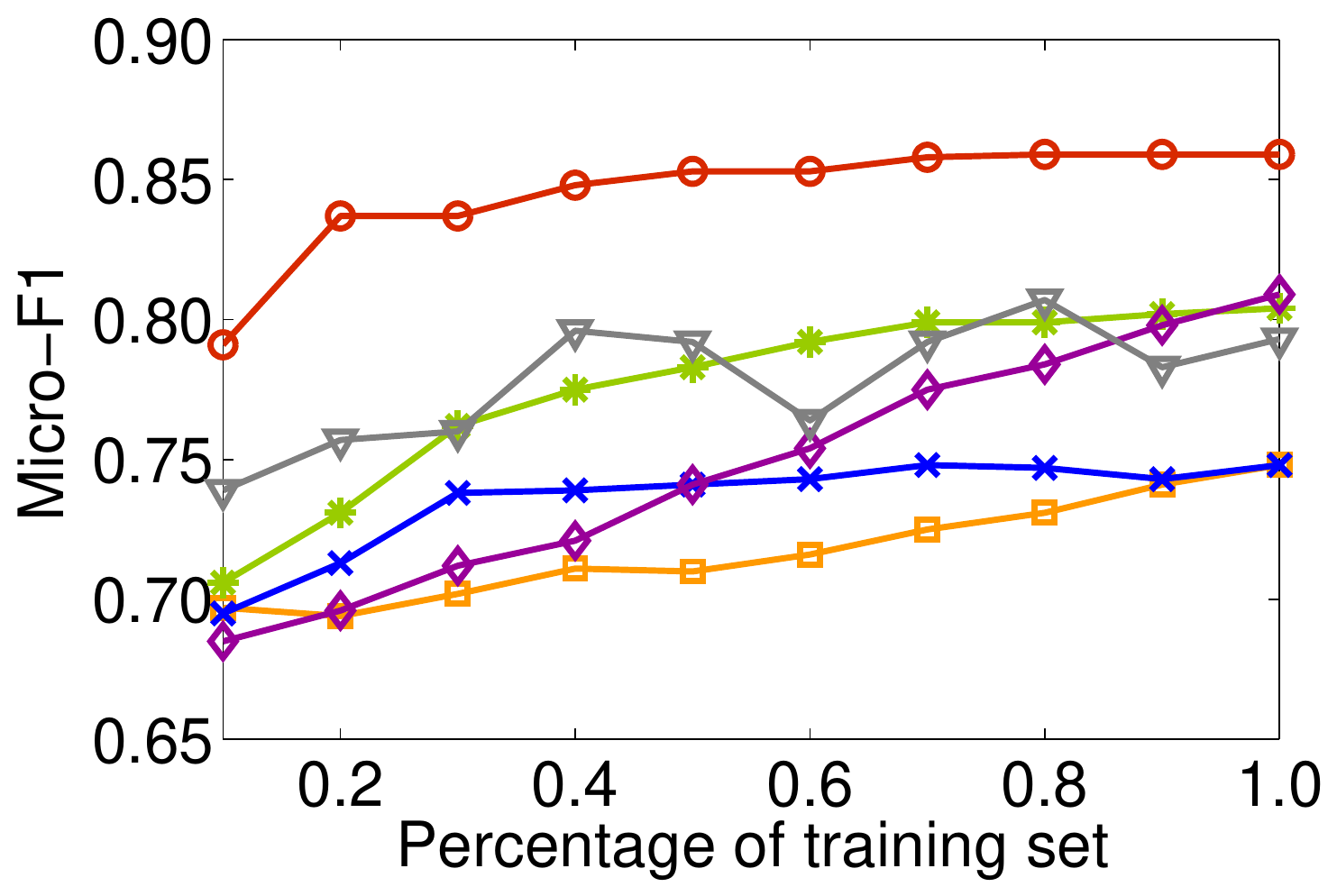}
				\vspace*{-0.1in}
				\caption{Micro-F1 on Wiki-RfA}
				\label{fig:5d}
			\end{subfigure}
  			\caption{Accuracy and micro-F1 on Weibo-STC and Wiki-RfA for link prediction.}
  			\label{fig:link_prediction}
		\end{figure*}
		
	\subsection{Parameter Setttings}
		We design a 4-layer autoencoder in SHINE for each network, in which the hidden layer is with 1,000 units and the embedding layer is with 100 units.
		Deeper architectures cannot further improve the performance but incur heavier computational overhead according to our experimental results.
		We choose concatenation as the aggregation function $g$ and inner product as the similarity measurement function $f$.
		Besides, we set the reconstruction weight of non-zero elements $\alpha = 10$, the balancing parameters $\lambda_1 = 1$, $\lambda_2 = 1$, $\lambda_3 = 20$, and $\lambda_4 = 0.01$ for SHINE.
		We will study the sensitivity of these parameters in Section \ref{sec:ps}.
		For LINE, we concatenate the first-order and second-order representations to form the final $100$-dimension embeddings for each node, and the total number of samples is $100$ million.
		For node2vec, the number of embedding dimension is set as $100$.
		For SDNE, the reconstruction weight of non-zero elements is $10$ and the weight of first-order term is $0.05$.
		For LIBFM, the dimensionality of the factorization machine is set as $\{1, 1, 0\}$ and we use SGD method for training with learning rate of 0.5 and 200 iterations.
		Other parameters in these baselines are set as default.
		
		In the following subsections, we conduct experiments on two tasks: link prediction and node recommendation.

	\subsection{Link Prediction}
		In link prediction setting, our task is to predict the sign of an unobserved link between two given nodes.
		As the existing links in the original network are known and can serve as the ground truth, we randomly hide $20\%$ of links in the sentiment network and select a balanced test set (i.e., the number of positive links is the same as negative links) out of them, while use the remaining network to train SHINE as well as all baselines.
		We use \textit{Accuracy} and \textit{Micro-F1} as the evaluation metrics in link prediction task.
		For a more fine-grained analysis, we compare the performance while varying the percentage of training set from $10\%$ to $100\%$.
		The result is presented in Fig. \ref{fig:link_prediction}, from which we have the following observations:
		\begin{itemize}
			\item
				Fig. \ref{fig:link_prediction} shows that our methods SHINE achieves significant improvements in Accuracy and Micro-F1 over the baselines in both datasets.
				Specifically, in Weibo-STC, SHINE outperforms LINE, node2vec, and SDNE by $13.8\%$, $16.2\%$, and $8.78\%$ respectively on Accuracy, and achieves $15.5\%$, $17.6\%$, $9.71\%$ gains respectively on Micro-F1.
			\item
				Among the three state-of-the-art network embedding methods, SDNE performs best while LINE and node2vec show relatively poor performance.
				Note that SDNE also uses autoencoder to learning the embedding of nodes, which proves the superiority of autoencoder in extracting highly nonlinear representations of networks from a side.
			\item
				FxG performs much better in Wiki-RfA than in Weibo-STC.
				This is probably due to the following two reasons:
				1) Unlike other methods, FxG cannot utilize the side information in Weibo-STC dataset.
				2) Weibo-STC is sparser than Wiki-RfA, which is unfavorable to the computing of goodness and fairness of nodes in FxG model.
			\item
				Although LIBFM is not specially designed for network-structured data, it still achieves fine performance compared with other network embedding methods.
				However, during experiments we find that LIBFM is unstable and prone to parameters tuning.
				This can also be validated by the fluctuating curves of LIBFM in Fig. \ref{fig:5c} and Fig. \ref{fig:5d}.
		\end{itemize}
		
		\begin{table}[t]
			\small
            \centering
            \caption{Comparison of models in terms of Accuracy and Micro-F1 on Weibo-STC in cold start scenario.}
            \begin{tabular}{|c|c|c|c|c|}
                \hline
                \multirow{2}{*}{Model} & \multicolumn{2}{c|}{Accuracy} & \multicolumn{2}{c|}{Micro-F1} \\
                \cline{2-5}
                & all users & new users & all users & new users \\
                \hline
                \textbf{SHINE} & \textbf{0.855} & \textbf{0.834} & \textbf{0.881} & \textbf{0.858} \\
                \hline
                \textbf{LINE} & 0.751 & 0.664 & 0.763 & 0.739 \\
                \hline
                \textbf{node2vec} & 0.736 & 0.653 & 0.749 & 0.667  \\
                \hline
                \textbf{SDNE} & 0.786 & 0.667 & 0.803 & 0.751 \\
                \hline
                \textbf{FxG} & 0.732 & 0.601 & 0.765 & 0.652 \\
                \hline
                \textbf{LIBFM} & 0.748 & 0.639 & 0.802 & 0.746 \\
                \hline
			\end{tabular}
			\label{table:cold_start}
		\end{table}
		
		To compare the performance of SHINE and baselines in cold start scenario, we construct a test set of newly arrived users for Weibo-STC, in which the associated ordinary user of each sentiment link dose not appear in the training set.
		We report Accuracy and Micro-F1 for all users and new users in Table \ref{table:cold_start}.
		From the results in Table \ref{table:cold_start} it is evident that SHINE can still maintain a decent performance in the cold start scenario, as it fully exploits the information from social network and profile network to compensate for the lack of sentiment links.
		By comparison, the performance of other baselines degrades significantly in cold start scenario.
		Specifically, the Accuracy decreases by $2.46\%$ for SHINE and by $11.58\%$, $11.28\%$, $15.14\%$, $17.90\%$, $14.57\%$ respectively for LINE, node2vec, SDNE, FxG and LIBFM, which proves that SHINE are more capable of effectively transferring knowledge among heterogeneous information networks, especially in cold start scenario.
		
	\subsection{Node Recommendation}
		\begin{figure*}[t]
			\centering
			\begin{subfigure}{.9\textwidth}
				\vspace*{-0.1in}
				\centering
				\includegraphics[width=\textwidth]{charts/legend.eps}
				\vspace*{-0.25in}
			\end{subfigure}
			\hfill
			\begin{subfigure}{.24\textwidth}
				\centering
				\includegraphics[width=\textwidth]{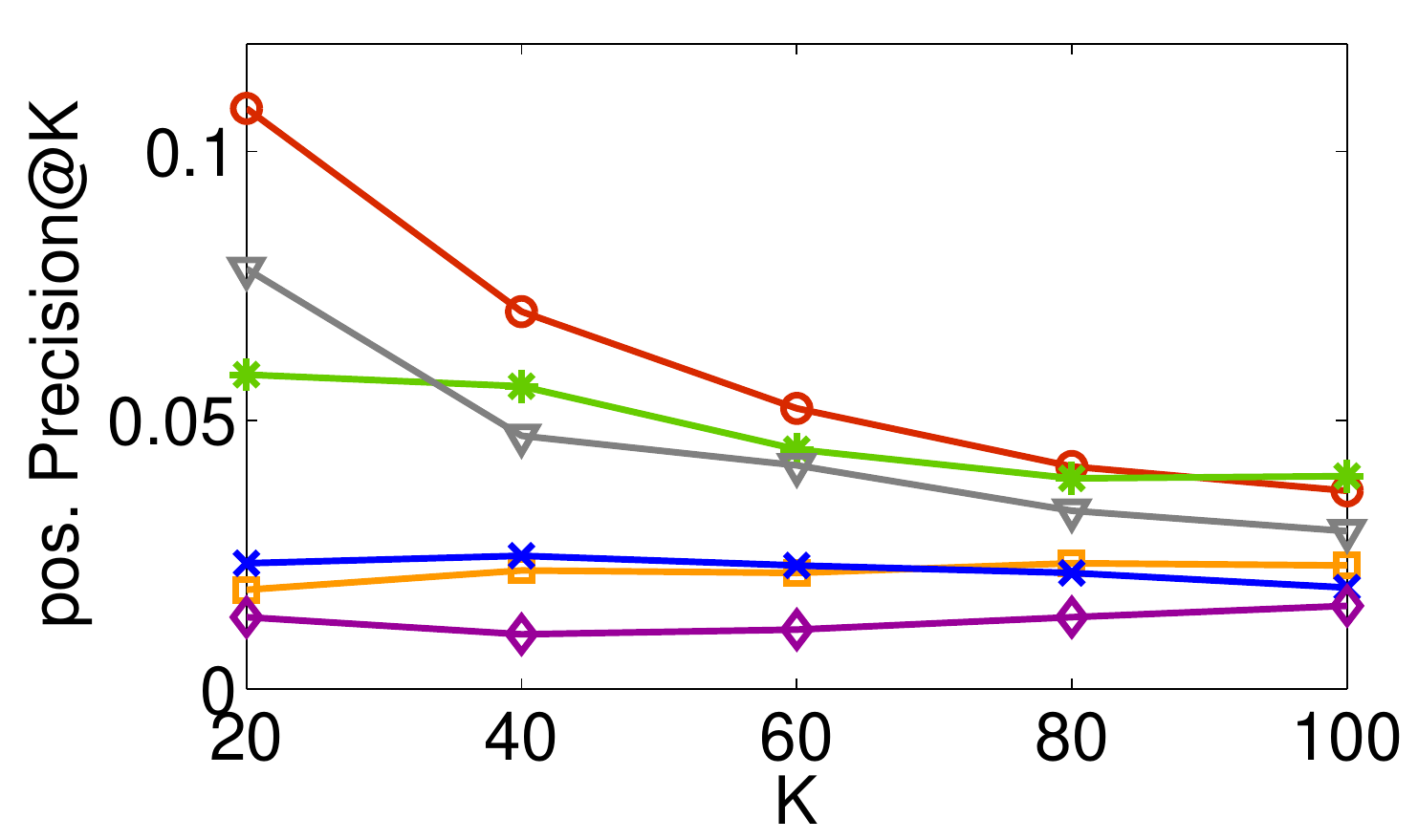}
				\vspace*{-0.2in}
				\caption{pos. Precision@K on Weibo-STC}
				\vspace{0.1in}
			\end{subfigure}
			\hfill
			\begin{subfigure}{.24\textwidth}
				\centering
				\includegraphics[width=\textwidth]{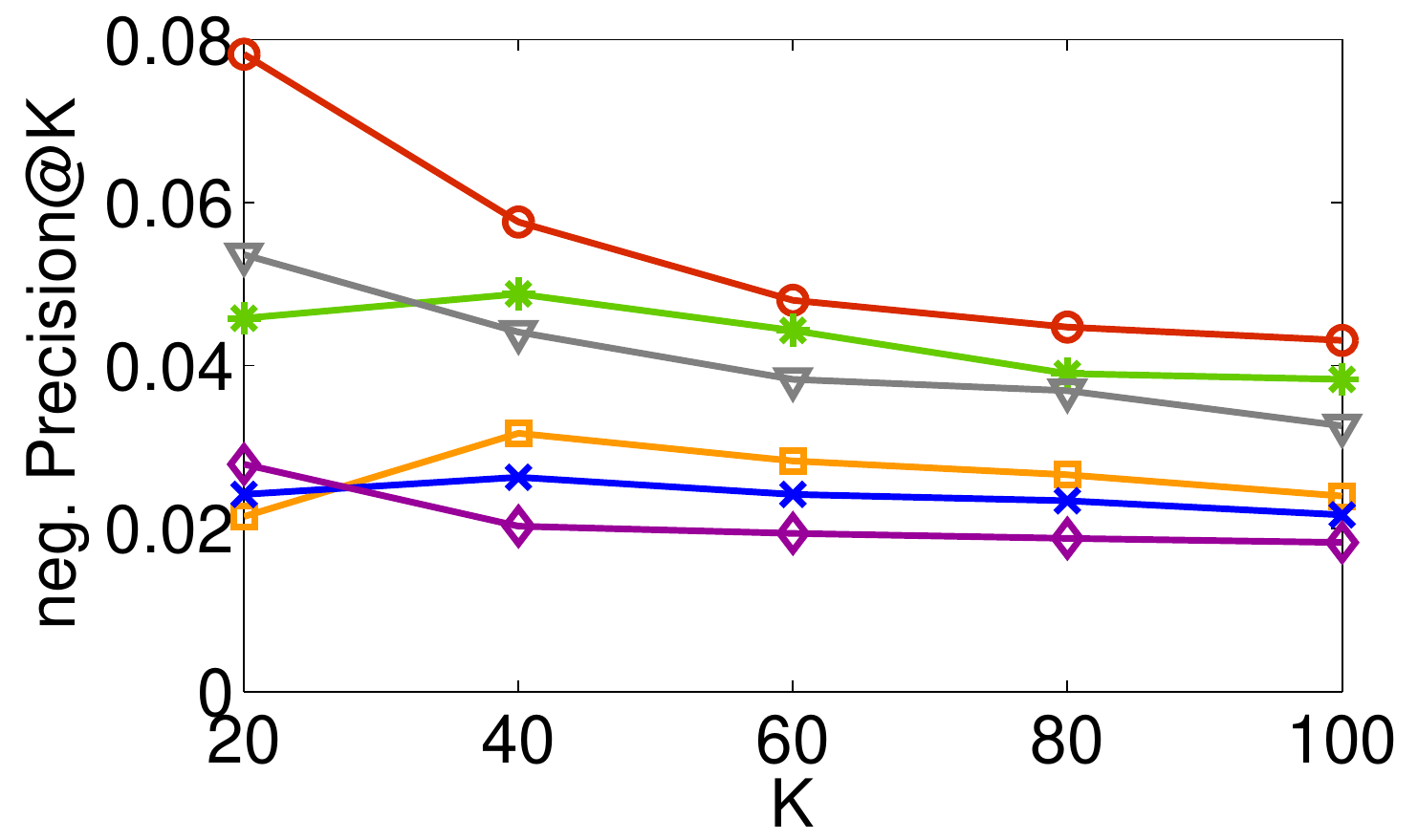}
				\vspace*{-0.2in}
				\caption{neg. Precision@K on Weibo-STC}
				\vspace{0.1in}
			\end{subfigure}
			\hfill
			\begin{subfigure}{.24\textwidth}
				\centering
				\includegraphics[width=\textwidth]{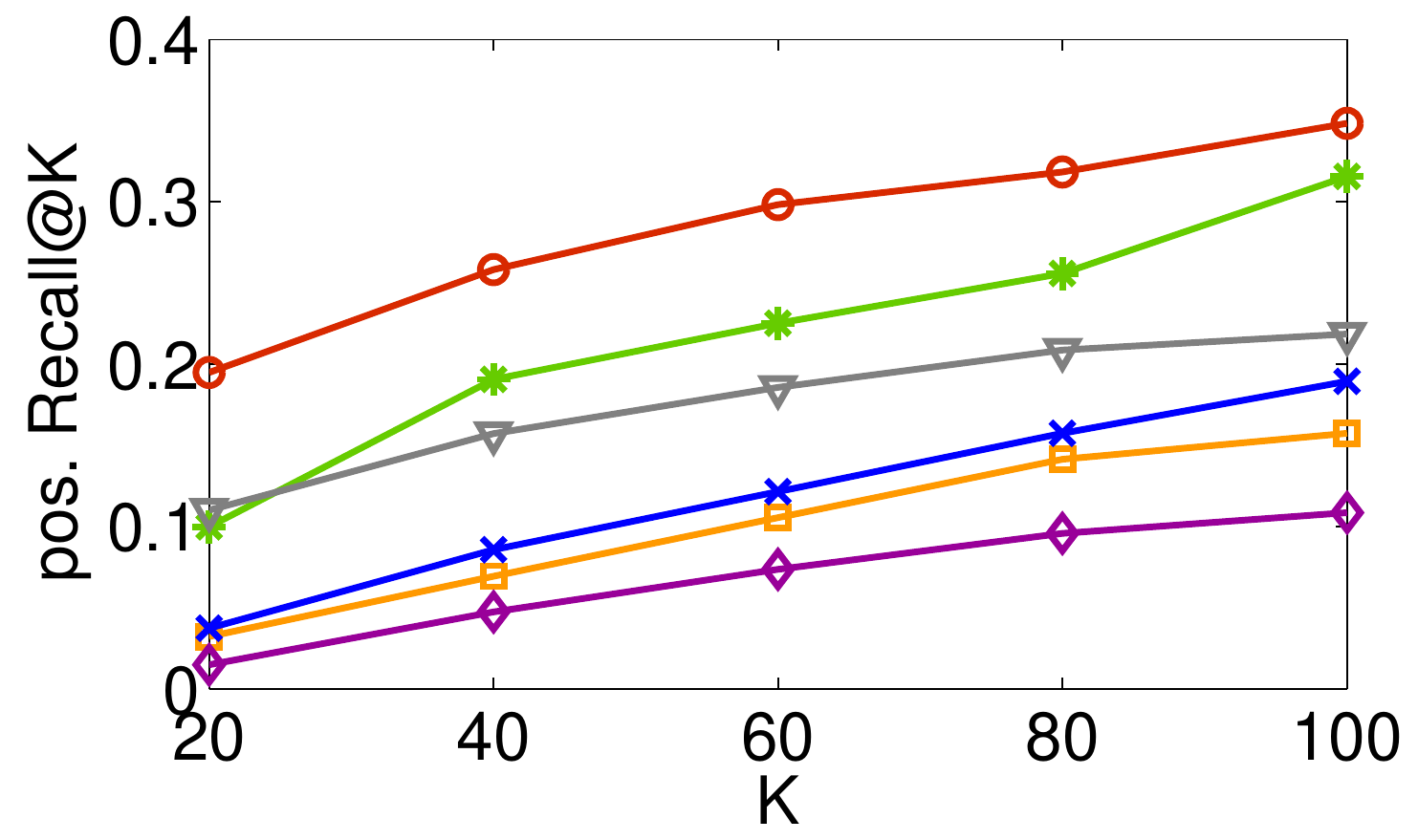}
				\vspace*{-0.2in}
				\caption{pos. Recall@K on Weibo-STC}
				\vspace{0.1in}
			\end{subfigure}
			\hfill
			\begin{subfigure}{.24\textwidth}
				\centering
				\includegraphics[width=\textwidth]{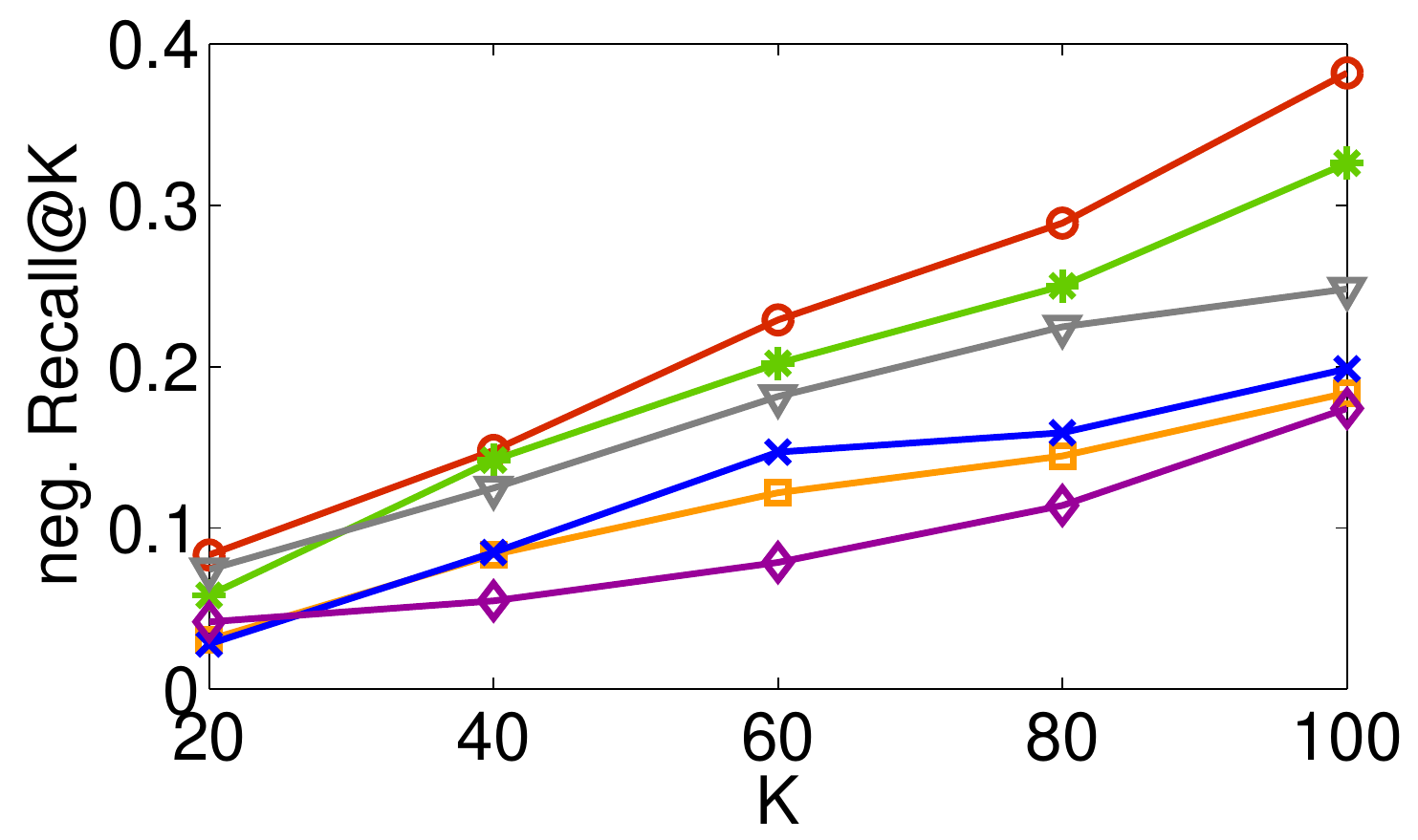}
				\vspace*{-0.2in}
				\caption{neg. Recall@K on Weibo-STC}
				\vspace{0.1in}
			\end{subfigure}
			\hfill
			\begin{subfigure}{.24\textwidth}
				\centering
				\includegraphics[width=\textwidth]{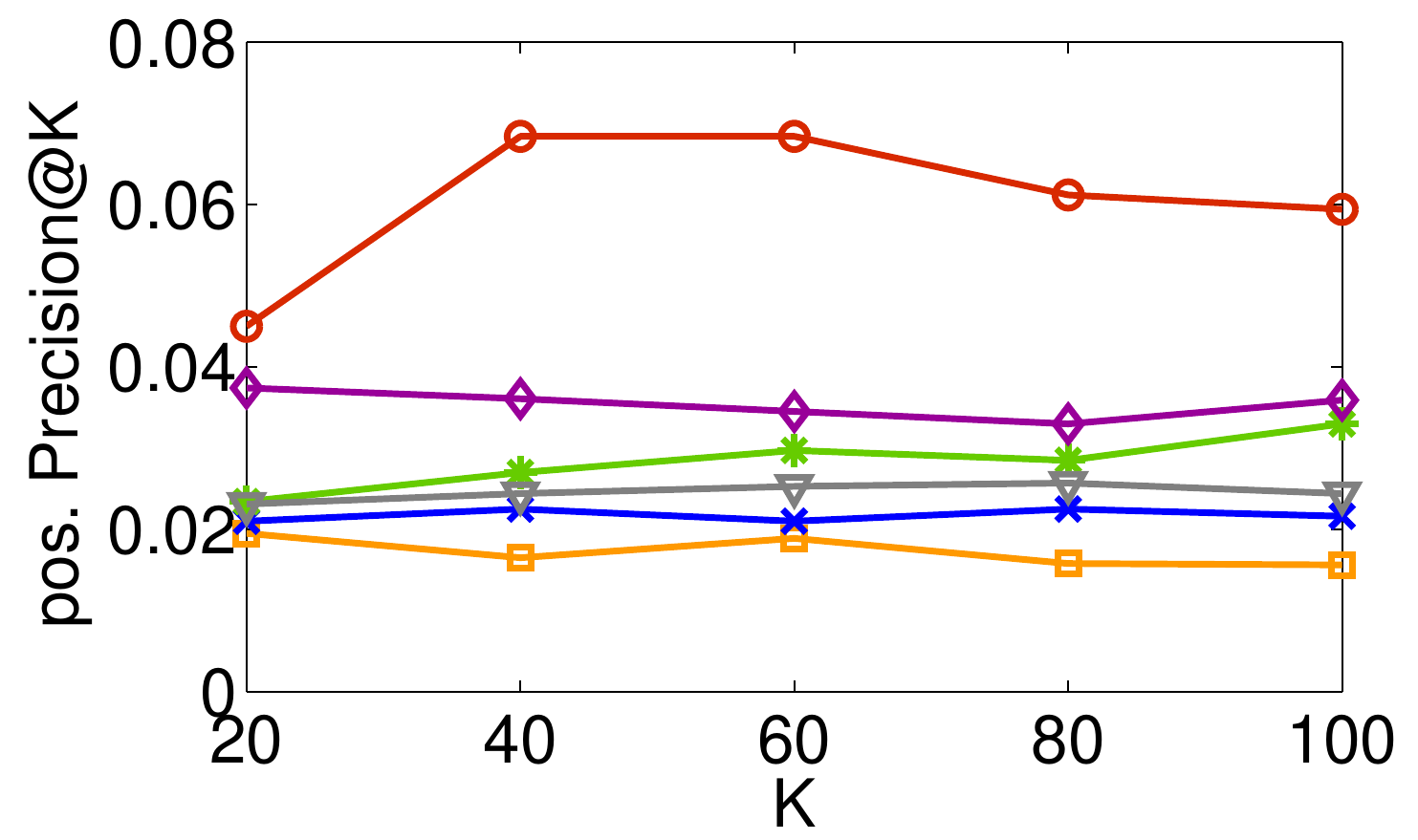}
				\vspace*{-0.2in}
				\caption{pos. Precision@K on Wiki-RfA}
				\vspace{0.1in}
			\end{subfigure}
			\hfill
			\begin{subfigure}{.24\textwidth}
				\centering
				\includegraphics[width=\textwidth]{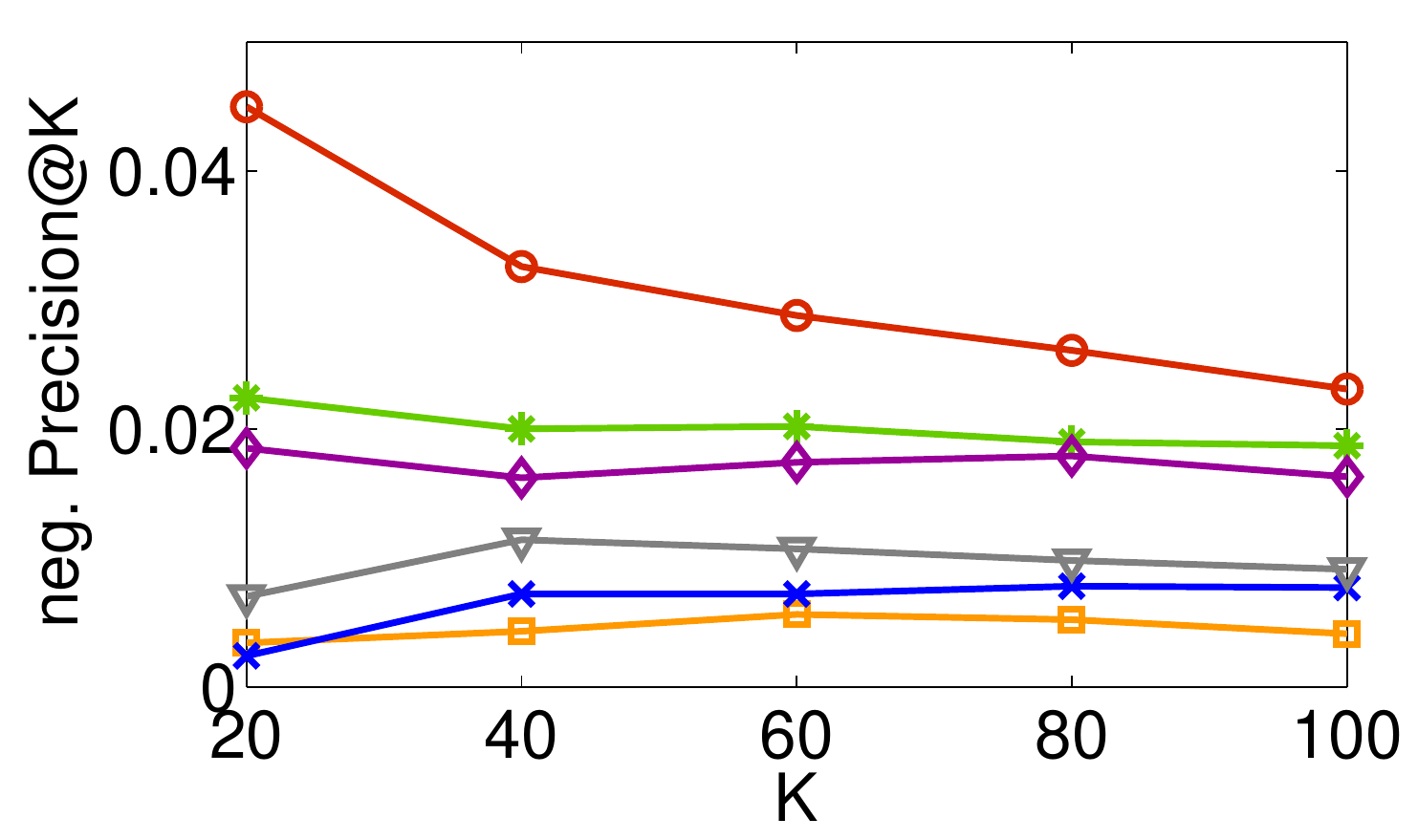}
				\vspace*{-0.2in}
				\caption{neg. Precision@K on Wiki-RfA}
				\vspace{0.1in}
			\end{subfigure}
			\hfill
			\begin{subfigure}{.24\textwidth}
				\centering
				\includegraphics[width=\textwidth]{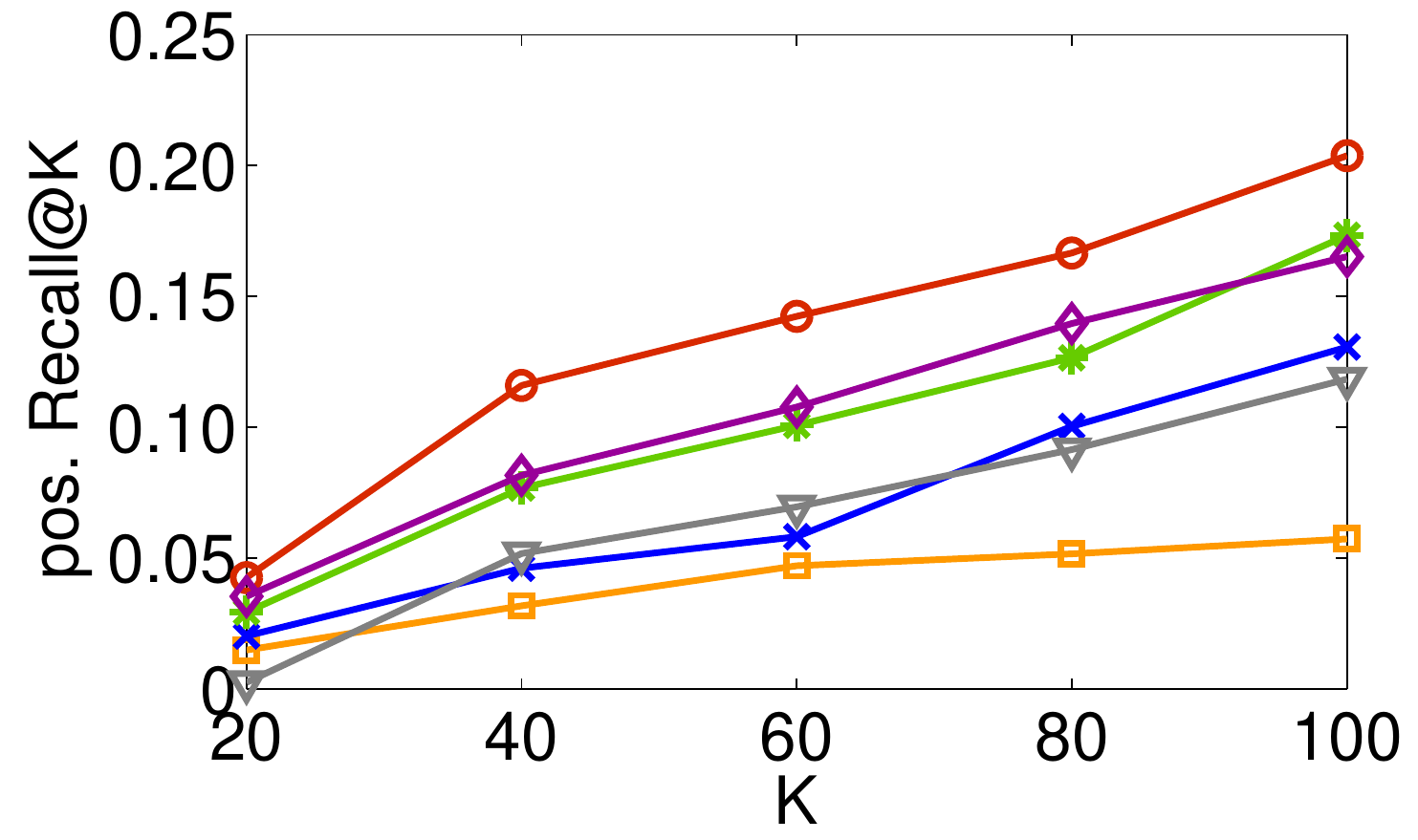}
				\vspace*{-0.2in}
				\caption{pos. Recall@K on Wiki-RfA}
				\vspace{0.1in}
			\end{subfigure}
			\hfill
			\begin{subfigure}{.24\textwidth}
				\centering
				\includegraphics[width=\textwidth]{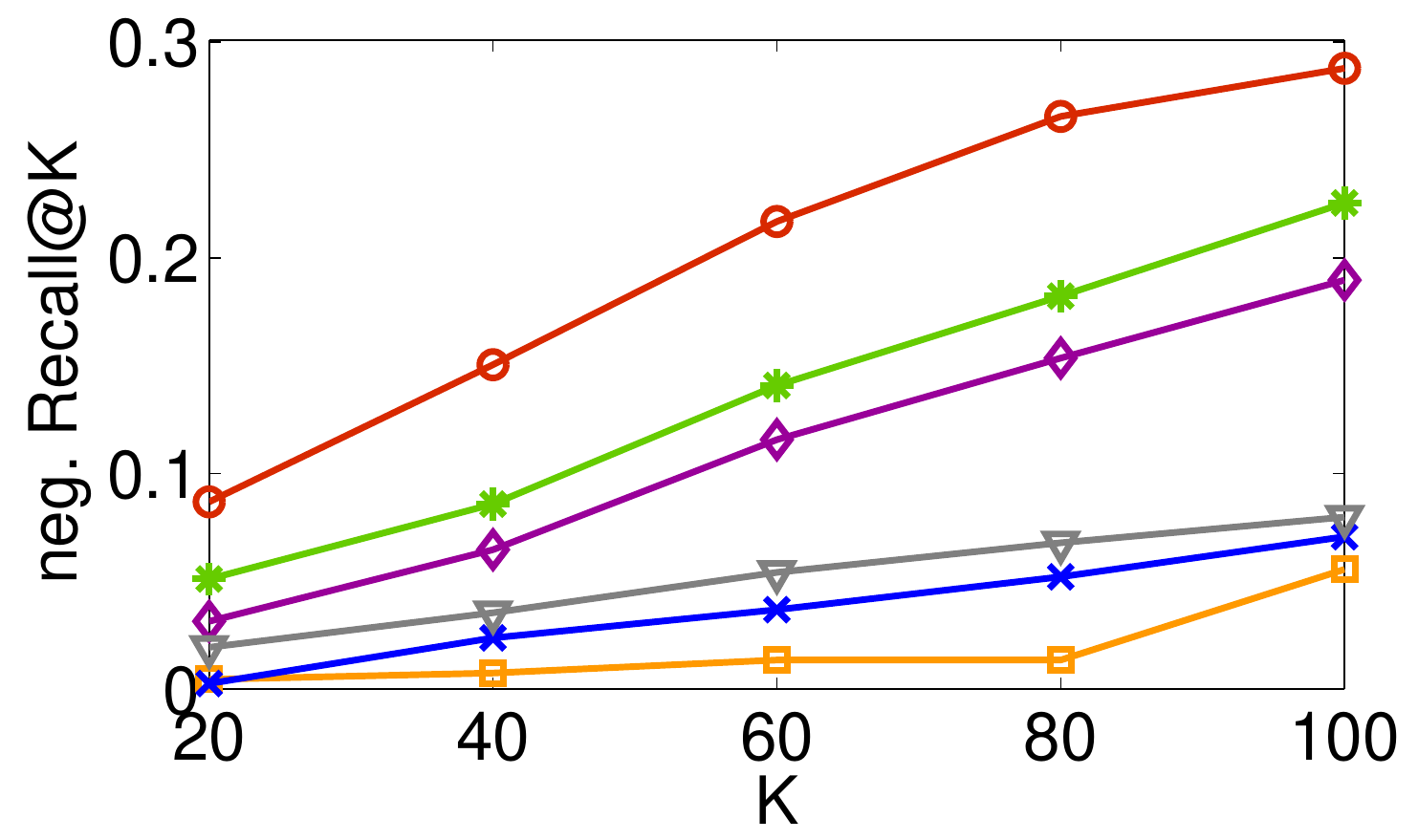}
				\vspace*{-0.2in}
				\caption{neg. Recall@K on Wiki-RfA}
				\vspace{0.1in}
			\end{subfigure}
  			\caption{Positive and negative Precision@K and Recall@K on Weibo-STC and Wiki-RfA for node recommendation.}
  			\label{fig:node_recommendation}
		\end{figure*}
	
		In addition to link prediction, we also conduct experiments on node recommendation, in which for each user we aim to recommend a set of users who have not been explicitly expressed attitude to but may be liked by the user.
		The performance of node recommendation can reveal the quality of learned representations as well.
		Specifically, for each user, we calculate his sentiment score toward all other users, and select $K$ users with largest sentiment score for recommendation.
		%Similar to the experiment in link prediction, we randomly hide $20\%$ of sentiment links as test set, and use Precision@K and Recall@K as the evaluation metrics for recommending new nodes.
		For completeness, we recommend not only the nodes that a user may like but also the nodes that he may dislike.
		Therefore, we use positive and negative Precision@K and Recall@K respectively for evaluation in corresponding experimental scenarios.
		The results are shown in Fig. \ref{fig:node_recommendation}, which provides us the following observations:
		\begin{itemize}
			\item
				The curve of SHINE is almost consistently above the curves of baselines, which proves that SHINE can better learn the representations of heterogeneous networks and perform recommendation than baselines.
			\item
				Negative precision is low than positive precision while negative recall is higher than positive recall for most methods.
				This is because negative links are far fewer than positive links in both datasets, which makes it easier to cover more negative links in the recommendation set.
			\item
				In general, the results of precision and recall on Weibo-STC is better than Wiki-RfA, which is in accordance with the results in link prediction.
				The reason lies in that Weibo-STC provides more side information which can greatly improve the quality of learned user representations.
		\end{itemize}

	\subsection{Parameters Sensitivity}\label{sec:ps}
		SHINE involves a number of hyper-parameters.
		In this subsection we examine how the different choices of parameters affect the Accuracy of SHINE on Weibo-STC dataset.
		Except for the parameter being tested, all other parameters are set as default.
			
		\begin{table}[t]
			\small
            \centering
            \caption{Accuracy on Weibo-STC w.r.t. the combinations of similarity measurement function and aggregate function.}
            \begin{tabular}{|c|c|c|c|}
                \hline
                \multirow{2}{*}{$f$} & \multicolumn{3}{c|}{$g$} \\
                \cline{2-4}
                & Summation & Max pooling & Concatenation\\
                \hline
                Inner product & 0.802 & 0.761 & \textbf{0.855}\\
                \hline
                Euclidean distance & 0.788 & 0.779 & 0.837\\
                \hline
                Logistic regression & 0.816 & 0.782 & 0.842\\
                \hline
			\end{tabular}
			\label{table:fg}
		\end{table}
		
		\textbf{Similarity measurement function $f$ and aggregation function $g$.}
		We first investigate how the similarity measurement function $f$ and aggregation function $g$ affect the performance by testing on all combinations of $f$ and $g$, and present the results in Table \ref{table:fg}.
		It is clear that the combination of inner product and concatenation achieves the best Accuracy, while max pooling performs worst, which is probably due to the reason that concatenation preserves more information out of the three types of embeddings than summation and max pooling during embedding aggregation.
		It should also be noted that there is no absolute advantage of all the three $f$ functions according to the results in Table \ref{table:fg}.

\begin{figure}[t]
			\centering
			\begin{subfigure}{.23\textwidth}
				\centering
				\includegraphics[width=\textwidth]{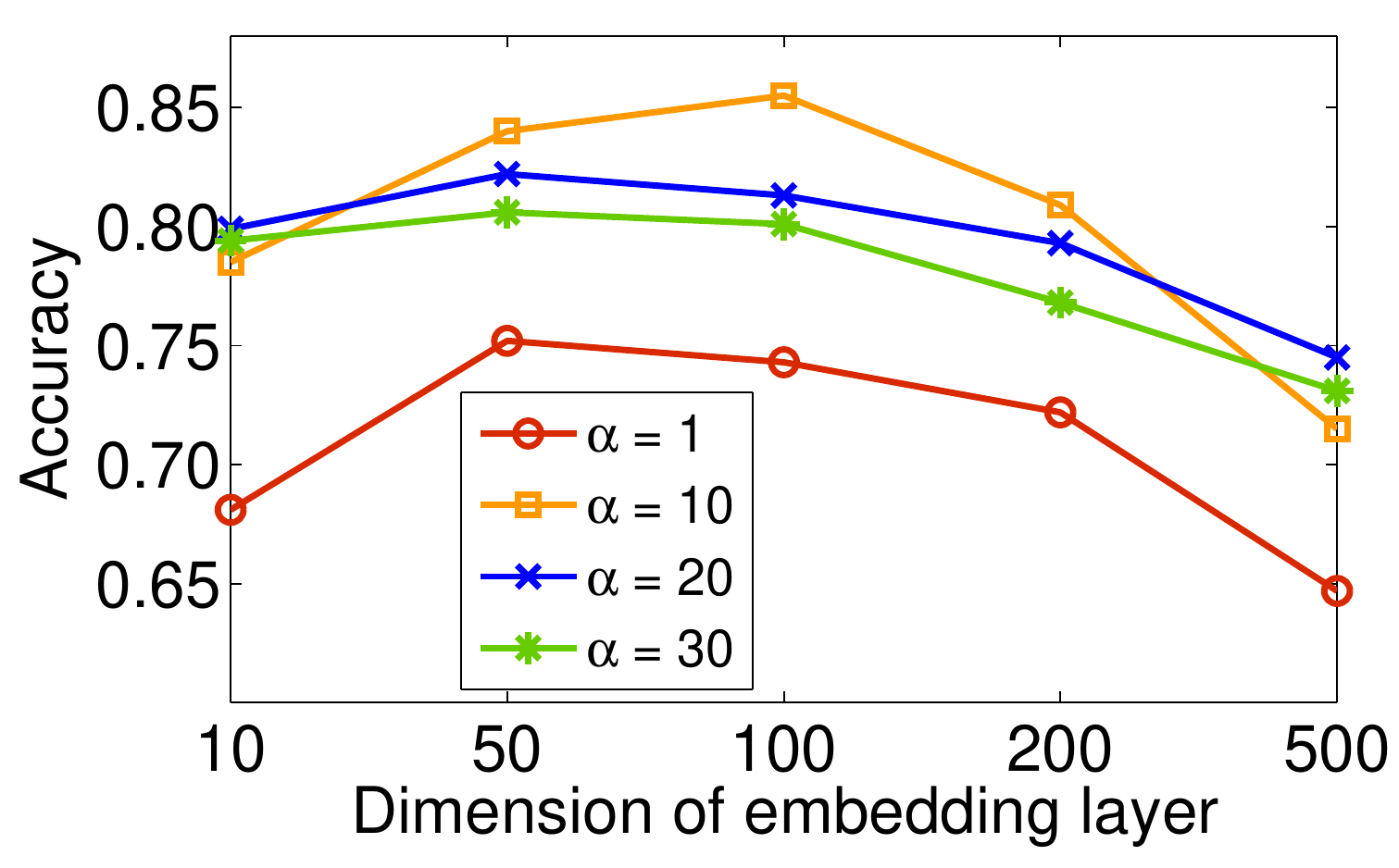}
				\vspace*{-0.2in}
				\caption{dim. of embedding layer and $\alpha$}
				\label{fig:ps_1}
			\end{subfigure}
			\hfill
			\begin{subfigure}{.23\textwidth}
				\centering
				\includegraphics[width=\textwidth]{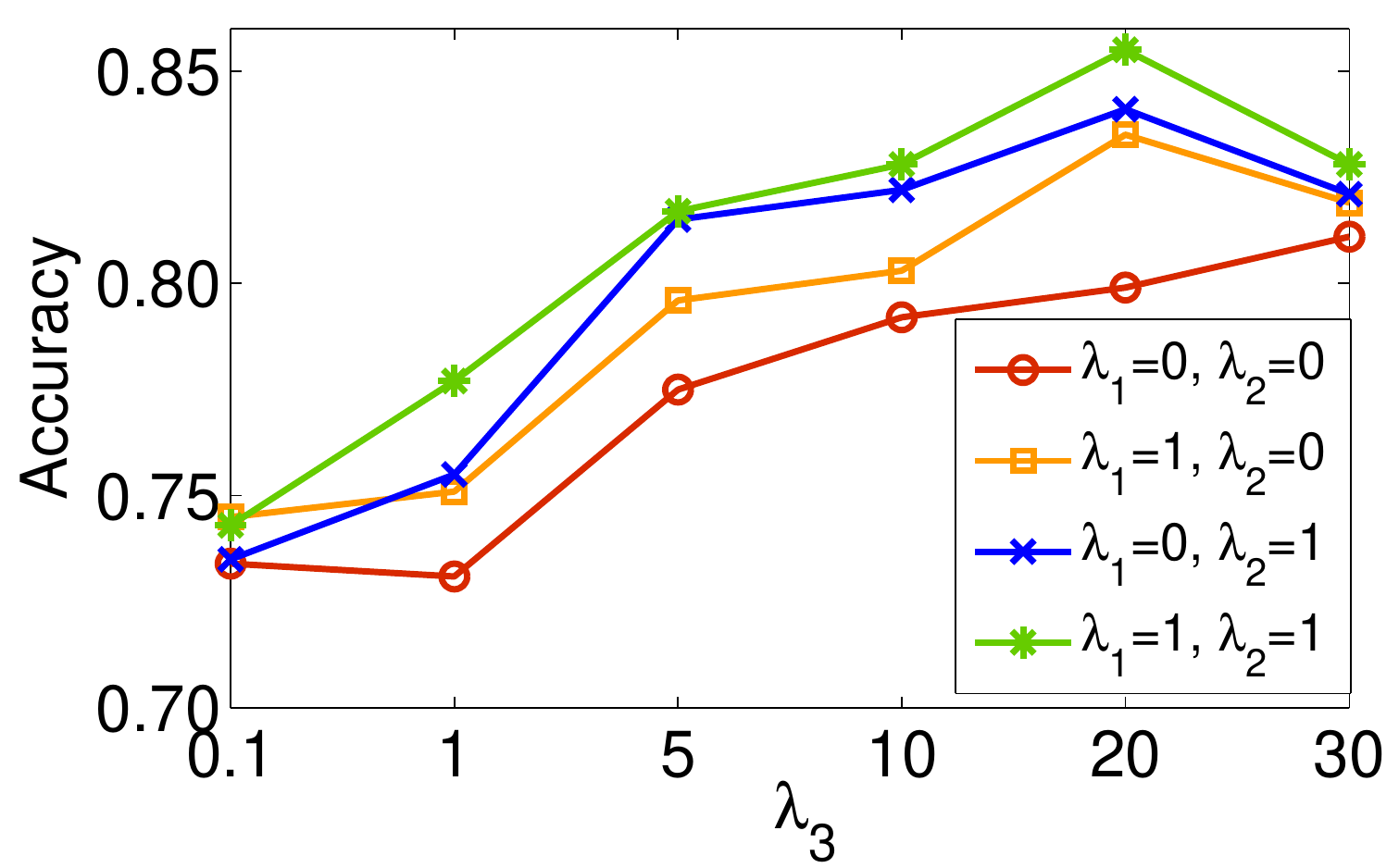}
				\vspace*{-0.2in}
				\caption{$\lambda_1$, $\lambda_2$, and $\lambda_3$}
				\label{fig:ps_2}
			\end{subfigure}
  			\caption{Parameter sensitivity w.r.t. the dimension of embedding layers, $\alpha$, $\lambda_1$, $\lambda_2$, and $\lambda_3$.}
  			\label{fig:ps}
		\end{figure}

		\textbf{Dimension of embedding layer and reconstruction weight of non-zero elements $\alpha$.}
		We also show how the dimension of embedding layer in the three autoencoders of SHINE and the hyper-parameter $\alpha$ affect the performance in Fig. \ref{fig:ps_1}.
		We have the following two observations:
		1) The performance is initially improved with the increase of dimension, because more bits in embedding layer can encode more useful information.
		However, the performance drops when the dimension further increases, as too large number of dimensions may introduce noises which mislead the subsequent prediction.
		2) $\alpha$ controls the reconstruction weight of non-zero elements in autoencoders.
		When $\alpha$ is too small (e.g., $\alpha = 1$), SHINE will reconstruct the zero and non-zero elements without much discrimination, which deteriorates the performance because non-zero elements are more informative than zero ones.
		However, the performance will decrease if $\alpha$ gets too large (e.g., $\alpha = 30$), because large $\alpha$ will lead SHINE to totally ignore the dissimilarity (i.e., zero elements) among users.
		
		\textbf{Balancing parameters $\lambda_1$, $\lambda_2$, and $\lambda_3$.}
		$\lambda_1$, $\lambda_2$, and $\lambda_3$ balance the loss terms of the objective function in Eq. (\ref{eq:objective_function}).
		We treat $\lambda_1$ and $\lambda_2$ as binary parameters and vary the value of $\lambda_3$ to study the performance of SHINE.
		Note that whether $\lambda_1$ or $\lambda_2$ equals 1 indicates that whether we use the additional social information or profile information in link prediction.
		Therefore, the study of $\lambda_1$ and $\lambda_2$ can also be seen as to validate the effectiveness of social network embedding module and profile network  embedding module.
		The result is presented in Fig. \ref{fig:ps_2}, from which we can conclude that:
		1) The curve of $\lambda_1 = 1, \lambda_2 = 0$ and $\lambda_1 = 0, \lambda_2 = 1$ are both above the curve of $\lambda_1 = 0, \lambda_2 = 0$, which demonstrates the significant gain by incorporating the social information and profile information (especially the latter) into the sentiment network.
		Moreover, combining both additional information can further improve the performance.
		2) Increasing the value of $\lambda_3$ can greatly boost the accuracy, as SHINE will concentrate more on the prediction error rather than the reconstruction error.
		However, similar to other hyper-parameters, too large $\lambda_3$ is not satisfactory since it breaks the trade-off among loss terms in objective function.
		
\section{Conclusions}\label{section_conclusions}
	In this paper we study the problem of predicting sentiment links in absence of sentiment related content in online social networks.
	We first establish a labeled, heterogeneous, and entity-level sentiment dataset from Weibo due to the lack of explicit sentiment links.
	To efficiently learn from these heterogeneous networks, we propose Signed Heterogeneous Information Network Embedding (SHINE), a deep-learning-based network embedding framework to extract users' highly nonlinear representations while preserving the structure of original networks.
	We conduct extensive experiments to evaluate the performance of SHINE.
	Experimental results prove the competitiveness of SHINE against several strong baselines and demonstrate the effectiveness of usage of social relation and profile information, especially in cold start scenario.

\begin{acks}
	We thank our anonymous reviewers for their feedback and suggestions.
	This work was partially sponsored by the National Basic Research 973 Program of China under Grant 2015CB352403.
\end{acks}

%\begin{acks}
%	The work is supported by ...
%\end{acks}

\bibliographystyle{ACM-Reference-Format}
\bibliography{sigproc} 

\end{document}